\documentclass[11pt]{article}
\usepackage[preprint]{acl}
\usepackage{times,latexsym,microtype,inconsolata}
\usepackage{pgfplots}
\pgfplotsset{compat=1.18}
\usepackage{graphicx}
\usepackage{subcaption,ragged2e}

\usepackage[normalem]{ulem}
\usepackage[utf8]{inputenc}
\usepackage[T1]{fontenc}
\usepackage{url}
\usepackage{xurl}
\usepackage{tikz}
\usepackage{enumitem}
\usetikzlibrary{positioning}
\usetikzlibrary{calc}
\usetikzlibrary{patterns}
\usetikzlibrary{shapes.geometric}
\usepgfplotslibrary{groupplots}
\usepackage{array}
\usepackage{booktabs}
\usepackage{wrapfig}
\usepackage{stfloats}
\usepackage{placeins}
\usepackage{caption}

\usepackage{listings}
\usepackage{adjustbox}
\usepackage{footnote}
\usepackage{multirow}
\usepackage{amsmath}
\usepackage{amsfonts,amssymb}
\usepackage{bbm}
\usepackage{xspace}
\usepackage{breakcites}
\usepackage{blindtext}
\usepackage{svg}
\usepackage[most]{tcolorbox}
\usepackage{tabularx}
\usepackage{graphicx}
\usepackage{xcolor}
\usepackage{float}
\usepackage{needspace}
\usepackage{makecell}
\usepackage{colortbl}
\newcommand{\newparagraph}[1]{\noindent\textbf{#1\hspace{0.5em}}}

\newcommand{\Ind}{
  \mathchoice
    {\scalebox{1.2}{$\displaystyle\mathbbm{1}$}}
    {\scalebox{1.2}{$\textstyle\mathbbm{1}$}}
    {\scalebox{1.1}{$\scriptstyle\mathbbm{1}$}}
    {\scalebox{1.0}{$\scriptscriptstyle\mathbbm{1}$}}
}

\newcommand{\acceptancelength}{AL\xspace}
\newcommand{\rift}{RIFT\xspace}
\newcommand{\riftstat}{RIFT-stat\xspace}
\newcommand{\riftdyn}{RIFT-dyn\xspace}
\newcommand{\riftada}{RIFT-ada\xspace}

\colorlet{reviewfixcolor}{black}
\DeclareRobustCommand{\reviewfix}[1]{#1}
\colorlet{clarificationcolor}{black}
\newif\ifnaaclrevisions
\naaclrevisionsfalse
\ifnaaclrevisions
  \definecolor{naacleditcolor}{RGB}{132,57,172}
\else
  \definecolor{naacleditcolor}{RGB}{0,0,0}
\fi
\colorlet{introrevision}{naacleditcolor}

\newcommand{\dt}[1]{}

\newcommand{\naaclPaperTitle}{RIFT: KV Cache Compression for Draft Models with Complementary Compressibility Training}
\title{\texorpdfstring{\protect\raisebox{-0.08em}{\protect\includegraphics[height=0.8em]{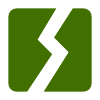}}\,}{ }\naaclPaperTitle}
\author{%
  \textbf{Dor~Tsur\NoHyper\thanks{Corresponding author: \href{mailto:dtsur@nvidia.com}{\texttt{dtsur@nvidia.com}}.}\endNoHyper\quad
  Maor~Ashkenazi\quad Maximilian~Golub\quad Peter~Hicks\quad Yaniv~Galron}\\[0.3em]
  \textbf{Talor~Abramovich\quad Yoav~Miron\quad Tiyasa~Mitra\quad
  Bita~Darvish~Rouhani\quad Ran~Zilberstein}\\[0.5em]
  \textnormal{NVIDIA}}
\begin{document}
\maketitle
\begin{abstract}
\begingroup
Speculative decoding is a popular decode acceleration method. 
As such, it is particularly relevant to long-output generation, which requires an increasingly large KV cache.
KV cache compression offers a way to constrain costs in long-context drafting. 
However, compression methods that focus on prefill information can lose relevance as the output length grows. We term this phenomenon \emph{decode staleness} and show that it degrades draft model accuracy under selection-based KV cache compression. We introduce \rift, a family of selection-based KV cache compression strategies that balance prefill information and recent generated tokens. We show that refreshing the KV cache selection during token generation significantly improves the drafter's quality at the cost of additional decode latency.
To alleviate the refresh costs, we propose an adaptive KV refresh strategy that reduces the number of score calculations while retaining the drafter's quality.
We further demonstrate how a compressibility-driven drafter training phase, based on sparse attention masks, can improve the drafter's quality under a compressed KV cache.
Combined with compressibility training, \rift enables Qwen3.5-4B, drafting for Qwen3.5-122B, to retain over 98\% of its full-cache acceptance length on 32K-token inputs, 4K-token outputs, and an 8K-token KV cache budget.
\par
\endgroup

\end{abstract}
\section{Introduction}
\label{sec:intro}
\definecolor{qoverviewnative}{RGB}{28,28,28}
\definecolor{qoverviewsw}{RGB}{135,135,135}
\definecolor{qoverviewsnap}{RGB}{205,164,0}
\definecolor{qoverviewreference}{RGB}{78,78,78}
\definecolor{qoverviewfixed}{RGB}{62,132,0}
\definecolor{qoverviewdynamic}{RGB}{0,145,105}
\definecolor{qoverviewadaptive}{RGB}{145,70,180}

\begingroup
\newcommand{\qsharedbar}[1]{
  \tikz[baseline=-0.5ex]{
    \filldraw[draw=#1,fill=#1!82,line width=0.4pt]
      (0,-0.06) rectangle (0.22,0.11);
  }
}
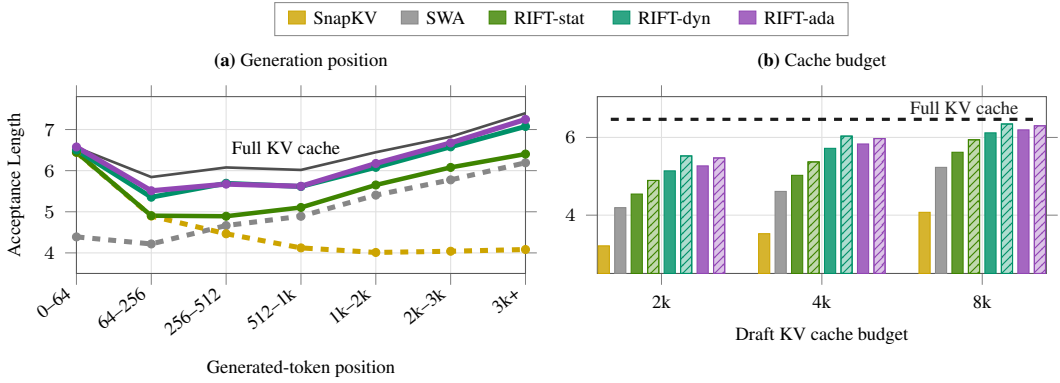
\begin{figure*}[t]
  \centering
  \begin{tikzpicture}
    \begin{groupplot}[
      group style={group size=2 by 1,horizontal sep=0.95cm},
      width=0.47\linewidth,
      height=0.245\linewidth,
      grid=major,
      grid style={black!12},
      tick align=outside,
      axis line style={black!65},
      label style={font=\scriptsize},
      tick label style={font=\scriptsize},
      title style={font=\scriptsize},
    ]
      \nextgroupplot[
        title={\textbf{(a)} Generation position},
        xmin=0,
        xmax=6,
        ymin=3.5,
        ymax=7.8,
        xtick={0,1,2,3,4,5,6},
        xticklabels={0--64,64--256,256--512,512--1k,1k--2k,2k--3k,3k+},
        x tick label style={rotate=38,anchor=east,font=\scriptsize},
        xlabel={Generated-token position},
        ylabel={Acceptance Length},
      ]
        \addplot[forget plot,color=qoverviewreference,mark=none,line width=1pt]
        coordinates {
          (0,6.564879) (1,5.844062) (2,6.077655) (3,6.017546)
          (4,6.447035) (5,6.824815) (6,7.396992)
        };
        \node[anchor=south,font=\scriptsize,text=black,inner sep=1pt]
          at (axis cs:2.8,6.32) {Full KV cache};
        \addplot[color=qoverviewsw,dashed,mark=*,mark size=1.5pt,mark options={solid,line width=0.4pt},line width=1.9pt]
          coordinates {
          (0,4.389059) (1,4.217744) (2,4.664816) (3,4.890495)
          (4,5.404534) (5,5.775898) (6,6.191576)
        };
        \addplot[color=qoverviewsnap,dashed,mark=*,mark size=1.5pt,mark options={solid,line width=0.4pt},line width=1.9pt]
          coordinates {
          (0,6.445432) (1,4.896680) (2,4.464991) (3,4.121568)
          (4,4.010871) (5,4.040032) (6,4.082093)
        };
        \addplot[color=qoverviewfixed,solid,mark=*,mark size=1.5pt,mark options={solid,line width=0.4pt},line width=1.7pt]
          coordinates {
          (0,6.447246) (1,4.904623) (2,4.891248) (3,5.104516)
          (4,5.651269) (5,6.078678) (6,6.403584)
        };
        \addplot[color=qoverviewdynamic,solid,mark=*,mark size=1.5pt,mark options={solid,line width=0.4pt},line width=1.85pt]
          coordinates {
          (0,6.503197) (1,5.354638) (2,5.692744) (3,5.614102)
          (4,6.080132) (5,6.578878) (6,7.074178)
        };
        \addplot[color=qoverviewadaptive,solid,mark=*,mark size=1.5pt,mark options={solid,line width=0.4pt},line width=1.95pt]
          coordinates {
          (0,6.575659) (1,5.510398) (2,5.672367) (3,5.620169)
          (4,6.174332) (5,6.674462) (6,7.245723)
        };

      \nextgroupplot[
        title={\textbf{(b)} Cache budget},
        ybar,
        bar width=4.2pt,
        ymin=2.5,
        ymax=7.05,
        symbolic x coords={2k,4k,8k},
        xtick=data,
        enlarge x limits=0.20,
        xlabel={Draft KV cache budget},
        ylabel={},
      ]
        \addplot[draw=qoverviewsnap,fill=qoverviewsnap!82] coordinates {
          (2k,3.210500) (4k,3.523300) (8k,4.073000)
        };
        \addplot[draw=qoverviewsw,fill=qoverviewsw!82] coordinates {
          (2k,4.193000) (4k,4.611000) (8k,5.230000)
        };
        \addplot[draw=qoverviewfixed,fill=qoverviewfixed!82] coordinates {
          (2k,4.543000) (4k,5.024800) (8k,5.618509)
        };
        \addplot[draw=qoverviewfixed,fill=qoverviewfixed!32,
          postaction={pattern=north east lines,pattern color=qoverviewfixed}]
          coordinates {
          (2k,4.893000) (4k,5.370000) (8k,5.939000)
        };
        \addplot[draw=qoverviewdynamic,fill=qoverviewdynamic!82] coordinates {
          (2k,5.139700) (4k,5.719500) (8k,6.118000)
        };
        \addplot[draw=qoverviewdynamic,fill=qoverviewdynamic!32,
          postaction={pattern=north east lines,pattern color=qoverviewdynamic}]
          coordinates {
          (2k,5.524000) (4k,6.034000) (8k,6.346000)
        };
        \addplot[draw=qoverviewadaptive,fill=qoverviewadaptive!82] coordinates {
          (2k,5.267982) (4k,5.831645) (8k,6.194716)
        };
        \addplot[draw=qoverviewadaptive,fill=qoverviewadaptive!32,
          postaction={pattern=north east lines,pattern color=qoverviewadaptive}]
          coordinates {
          (2k,5.472187) (4k,5.970717) (8k,6.299506)
        };

        \draw[color=qoverviewnative,dashed,line width=1.2pt]
          (rel axis cs:0.03,0.87165) -- (rel axis cs:0.97,0.87165);
        \node[anchor=south east,font=\scriptsize,text=qoverviewnative,inner sep=1pt]
          at (rel axis cs:0.945,0.884) {Full KV cache};
    \end{groupplot}
    \node[anchor=south,draw=black!45,line width=0.4pt,fill=white,
      inner xsep=5pt,inner ysep=3pt,font=\scriptsize] at
      ($(group c1r1.north west)!0.5!(group c2r1.north east)+(0,0.80cm)$)
      {\qsharedbar{qoverviewsnap}\,SnapKV\quad
       \qsharedbar{qoverviewsw}\,SWA\quad
       \qsharedbar{qoverviewfixed}\,\riftstat\quad
       \qsharedbar{qoverviewdynamic}\,\riftdyn\quad
       \qsharedbar{qoverviewadaptive}\,\riftada};
  \end{tikzpicture}
  \caption{AL improvement of \rift and compressibility training on a Qwen3.5-4B draft (Qwen3.5-122B target), SPEED-Bench 32K ISL low-entropy, draft length 11.
  \textbf{(a)} AL vs. generation interval at 8K cache budget.
  \textbf{(b)} AL vs. KV cache budget. AL is measured vs. the full-KV draft AL. Diagonal pattern denotes compression-aware training.
  }
  \label{fig:qwen35-overview}
  \label{fig:qwen35-overview-shared-legend}
\end{figure*}
\endgroup
\newcommand{\introedit}[1]{{\color{introrevision}\hypersetup{citecolor=introrevision,linkcolor=introrevision,urlcolor=introrevision}#1}}

Speculative decoding accelerates autoregressive language models by verifying multiple draft proposals in a single target-model pass while preserving the target distribution~\cite{leviathan2023fast,chen2023accelerating}, shifting inference towards a compute-bound regime.
Its effectiveness depends on a draft that is both inexpensive and closely aligned with the target. 
Reasoning and agentic applications increase the pressure on this balance, combining long input sequence lengths (ISLs) with long output sequence lengths (OSLs) as models generate reasoning traces and accumulate interaction history~\cite{nye2021show,yao2023react}. 
As context grows, key-value (KV) cache access
can increasingly bottleneck drafting, limiting the achievable speedup from speculation~\cite{sadhukhan2025magicdec}.
As a decode acceleration method, speculative decoding targets long OSL, further magnifying the KV cache bottleneck.
This bottleneck is still evident in hybrid attention architectures~\cite{dao2024transformersssmsgeneralizedmodels,yang2024gateddeltanetworks}: even when only 25\% of a small hybrid model's layers use full attention, its KV footprint can significantly exceed its weights' memory footprint at long contexts with only a few concurrent requests (See Appendix~\ref{app:draft-kv-memory} for details).
Existing literature considers various forms of KV compression~\cite{hooper2024kvquant,liu2024kivi,laus2026kv}. Here, we focus on selection-based compression, where decode tokens attend only a compact subset of their past KVs.

Tail-augmented drafters, e.g., MTP, Eagle or DFlash~\cite{li2026eagle,chen2026dflash}, are considered superior to external draft models on GPU-based deployments due to their minimal latency overhead. However, contemporary hardware-disaggregated inference and edge-assisted serving renews the relevance of small external drafts~\cite{aubrey2026groq3lpx,park2025specedge} over tail-augmented blocks, such as MTPs or DFlash~\cite{chen2026dflash}. In these settings, memory capacity poses a constraint, as memory is shared between the model weights and growing KV cache~\cite{kim2026memspec}.
Considering all the above, we study draft KV cache compression under a fixed cache budget that is ISL-independent.
KV compression also enables KV-based self-speculation: the target model itself drafts with a compressed cache and verifies with full cache~\cite{sun2024triforce,sadhukhan2025magicdec,zhao2025sparsespec}.
Together, these regimes motivate the question:
\begin{center}
\emph{How can we best compress the drafter's KV cache to sustain target alignment in long contexts?}

\end{center}

We study this question through the trade-off between acceptance length (AL) and the KV cache budget size, measuring drafting quality under a fixed memory constraint.
A fixed KV-budget introduces a tradeoff between KVs selected during prefill and KVs accumulated during decode. Prompt tokens can provide valuable information, while the generated prefix records the target's recent context. A prefill-based selection becomes stale as generation evolves, while a sliding window (SW) approach limits the model's view to the latest prefix tokens and discards remote information. 
Decode-time refresh of selected KVs allows to \reviewfix{adaptively} determine the relevant KVs, at the cost of additional latency.
This motivates two complementary approaches: allocate capacity to both prefill positions and generated tokens, and refresh the selected history as relevance changes.

Our analysis characterizes \emph{decode staleness} and introduce
\textbf{\rift} (\textbf{R}ecency--\textbf{I}mportance \textbf{F}usion and \textbf{T}raining), a family of selection-based KV compression strategies combining recency allocation with decode-time refresh of importance-selected context. We further propose compression-aware knowledge distillation (KD) to reduce the training–inference mismatch and improve draft alignment under compression.
Our contributions are:
\begin{enumerate}[leftmargin=*,itemsep=3pt,topsep=3pt]
    \item \textbf{Characterizing decode staleness:} We show how prefill selection loses its initial advantage over SWA as generation progresses, and show this gap increases in hybrid-attention models.
    \item \textbf{\rift:} We address decode staleness and propose \rift, a family of KV compression methods for draft models, balancing implementation complexity with resulting drafting quality. 
    \item \textbf{Draft-compressibility training:} We develop a knowledge-distillation scheme with compression-driven attention masks to improve draft alignment under a restricted KV budget. 
\end{enumerate}
Together, our methods retain most of the draft's acceptance length, while using a fraction of the total KV cache. In SPEED-Bench~\cite{speedbenchnvidia} with 32K ISL and $4\times$ compression ratio, Qwen3.5-4B retains up to $~96\%$ of its full cache AL (Qwen3.5-122B target), and Nemotron3.5-Lightning retains $~97\%$ of its full-cache AL (Nemotron3-Ultra target), while SnapKV retains $~63-67\%$ AL in both cases.

\section{Background and Setting}
\label{sec:background}

\textbf{Scope.}
We focus on two settings: 1) external autoregressive draft models and 2) KV-based self-speculative decoding. We leave tail-augmented drafters outside the scope of this work, as our preliminary studies find negligible benefit from KV compression in these settings (Appendix~\ref{app:tail-augmented-kv}).

\textbf{Draft-side Compression.} 
Rejection sampling preserves the target distribution regardless of the draft's KV-compression policy. Since compressing the target cache can instead degrade output quality, we compress only the draft cache. We measure accuracy using average acceptance length (AL), the expected number of accepted tokens plus the extra target-generated token.
We consider two strategies representing the endpoints of the spectrum of static KV selection methods under a budget $B$: 

\textbf{SW attention.}
Pure SW attention (SWA) retains $B$ recent
tokens, updating the cache at every decode step, independently of token importance.

\textbf{Prefill-based KV compression.}
Prefill-based methods select KVs using importance scores
aggregated during prefill. A widely used method,
SnapKV~\cite{li2024snapkv}, aggregates attention scores from a set of last prompt queries, termed the
\emph{observation window}. For each KV head, SnapKV retains the highest-scoring KV positions alongside a SW the size of the observation window. Herein, we take an observation window of $64$ tokens.

\section{Decode Staleness In KV Compression}
\label{sec:root-cause}
\colorlet{rcarevision}{naacleditcolor}
\newcommand{\rcaedit}[1]{\textcolor{rcarevision}{#1}}

While prefill-only KV compression has proven beneficial for speculative decoding~\cite{sadhukhan2025magicdec}, prior evaluations do not fully reflect long OSLs and modern model architectures.
We start by investigating prefill-based methods in speculative decoding. We use SnapKV to compress the KV cache of Qwen3.5-4B, drafting for Qwen3.5-122B and evaluate on SPEED-Bench 32K ISL low entropy split~\cite{speedbenchnvidia}, with reasoning enabled.
As seen in Figure~\ref{fig:qwen35-overview}(b), SWA outperforms SnapKV given the same KV budget.
This result is surprising: SnapKV incorporates importance information, while SWA naively keeps a fixed tail of previous tokens. We make the following observations about the KV-compression literature:

\textbf{KV cache compression benchmarks often overlook long OSL.}
Evaluations centered on long inputs primarily test which prompt KVs to retain~\cite{bai2024longbench,hsieh2024ruler}. Regarding decode tokens, \reviewfix{one can} either 1) retain all generated KVs, 2) maintain a small sliding window alongside the prefill-selected cache, or 3) rescore and update the selection at \textit{every} decoding step. The tradeoffs between these options become particularly important in speculative decoding, where the draft must remain aligned with a growing target-verified prefix under a fixed KV budget. Retaining all generated KVs \reviewfix{violates the budget} constraint, while a small sliding window can compromise alignment later in generation (as seen in Figure~\ref{fig:qwen35-overview}(a)). Rescoring at every decode step adds significant overhead to the drafting process.

To understand why a fixed prefill selection becomes less effective during generation, we examine how its relevance degrades, a phenomenon we term \emph{decode staleness}, which motivates the following hypothesis:

\begin{tcolorbox}[
  colback=blue!2,
  colframe=blue!35!black,
  coltext=black,
  boxrule=0.5pt,
  arc=1mm,
  left=5pt,right=5pt,top=4pt,bottom=3pt,
  before skip=7pt,after skip=9pt]
  \textbf{Decode staleness:} prefill-based KV selections become stale as generation progresses.
\end{tcolorbox}

\noindent Additionally, we note that \textit{KV cache compression methods are rarely evaluated on hybrid-attention models}. Yet, contemporary models interleave full attention with efficient recurrent components, including linear attention in Qwen3.5~\cite{yang2024gateddeltanetworks,qwen2026qwen35} and Mamba layers in Nemotron 3~\cite{dao2024transformersssmsgeneralizedmodels,nvidia2026nemotron3superopen}.

\textbf{Output-position dependence.}
We analyze average draft AL across generation positions with an $8$K-token budget ($4\times$ compression).
As shown in Figure~\ref{fig:qwen35-overview}(a), SnapKV's AL decreases by 36.7\% while SW improves by 41\% as generation evolves.
We term the evolving SnapKV-SW difference the \textit{staleness gap}, which we consider a performance-based proxy for the growing irrelevance of SnapKV's selected KVs during generation.
We attribute the growth of the staleness gap to two factors. First, in speculative decoding the recent decode prefix is produced by the target model, making access to it highly important for draft-target alignment. Second, recent decode tokens often contain reasoning traces, which potentially act as a running prompt summary~\cite{nye2021show,yao2023react}. Indeed, preliminary study shows that disabling reasoning reduces the staleness gap from $2.11$ to $0.60$ AL (See Figure~\ref{fig:rca-qwen-thinking-position} in Appendix~\ref{app:root-cause}).

\textbf{Architecture dependence.}
We empirically observe that decode staleness is exacerbated in hybrid-attention architectures, which are often overlooked in the KV compression literature.
Due to lack of space, we refer the reader to Appendix~\ref{app:root-cause}, where we show that 1) a full-attention Llama3.1 draft retains the expected ordering: SnapKV consistently outperforms SWA throughout generation, while, 2) in Nemotron3 (which is hybrid-SSM) and GPT-OSS (which interleaves hard-coded SWA layers with full attention), SWA again outperforms SnapKV.

\textbf{Mechanistic evidence.}
We measure the fraction of the full-cache draft's attention assigned to positions retained by each policy (SnapKV and SWA), under an 8K budget.
Figure~\ref{fig:rca-attention-allocation} shows that this fraction drops from $67.0\%$ to $25.4\%$ for SnapKV, while increasing from $43.5\%$ to $68.7\%$ for SWA as generation progresses.
This crossover is consistent with attention shifting toward generated context that the prefill-selected cache largely omits.

\definecolor{rcaprompt}{RGB}{76,114,10}
\definecolor{rcagenerated}{RGB}{85,10,10}
\definecolor{rcasw}{RGB}{31,119,180}
\definecolor{rcasnap}{RGB}{214,79,55}

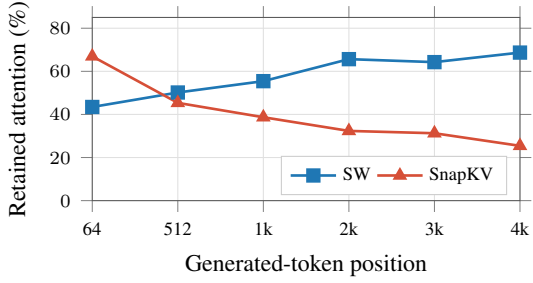
\begin{figure}[t]
  \centering
  \begin{tikzpicture}
    \begin{groupplot}[
      group style={group size=1 by 1,vertical sep=1.35cm},
      width=0.94\linewidth,
      height=0.52\linewidth,
      xmin=0,
      xmax=5,
      ymin=0,
      ymax=85,
      xtick={0,1,2,3,4,5},
      xticklabels={64,512,1k,2k,3k,4k},
      xlabel={Generated-token position},
      grid=major,
      grid style={black!12},
      tick align=outside,
      axis line style={black!65},
      every axis plot/.append style={line width=1pt,mark size=2.2pt},
      label style={font=\footnotesize},
      tick label style={font=\scriptsize},
      title style={font=\footnotesize},
      legend style={draw=black!20,fill=white,font=\scriptsize,
        at={(0.97,0.03)},anchor=south east,legend columns=2},
    ]
      \nextgroupplot[ylabel={Retained attention (\%)}]
        \addplot[color=rcasw,mark=square*] coordinates {
          (0,43.4537) (1,50.1706) (2,55.4399)
          (3,65.6515) (4,64.2667) (5,68.6939)
        };
        \addlegendentry{SW}
        \addplot[color=rcasnap,mark=triangle*] coordinates {
          (0,66.9504) (1,45.3663) (2,38.6963)
          (3,32.3477) (4,31.2686) (5,25.4263)
        };
        \addlegendentry{SnapKV}
    \end{groupplot}
  \end{tikzpicture}
  \caption{Fraction of full-KV attention mass retained by SW and SnapKV (with 64 token SW) in the Qwen3.5-4B draft on 32K-ISL data, with an 8K budget. SnapKV retains less attention as generation proceeds, while SW retains more.}
  \label{fig:rca-attention-allocation}
\end{figure}

\section{\rift: Staleness-Aware KV cache Compression}
\label{sec:staleness-aware-kv-compression}
Motivated by the staleness gap, in this section we propose \rift, a family of KV compression policies for draft models. \rift comprises several solutions, ranging by their implementation complexity.

\subsection{Effect of Sliding Window Ratio on Prefill-Based Methods}
\label{sec:sliding-window-ratio}

Our analysis emphasizes the importance of recent tokens to the draft's KV cache.
Thus, our first solution, which we call \riftstat, is a simple modification to a given static KV selection method.\footnote{We call a KV selection strategy \textit{static} if it does not change during decode.}
\riftstat balances a given \reviewfix{budget} $B$ between a \reviewfix{prefill-based token} selection \reviewfix{algorithm} (e.g. SnapKV) and a SW component. For a given SW ratio $\rho\in(0,1]$ and a cache budget $B$, we reserve $r=\lfloor\rho B\rfloor$ positions for the most recent tokens and use the remaining $k=B-r$ positions for attention-based KV allocation.
The SW ratio $\rho$ controls the trade-off between long-range retrieval and recent streaming context.
Figure~\ref{fig:effect-sw-ratio} shows the draft's \acceptancelength as a function of $\rho$ combining \riftstat with SnapKV (Section~\ref{sec:results} presents results on several KV selection methods).
With an $8K$ budget, the highest \acceptancelength is achieved at $\rho=0.5$ (AL $5.614$), an improvement of $37.8\%$ over SnapKV (AL $4.073$), with pure SWA \reviewfix{achieving} $5.230$. In Appendix~\ref{app:sw-ratio-additional} we provide additional analysis on the effect of the SW ratio in \riftstat.

\definecolor{swratioLine}{RGB}{31,119,180}
\definecolor{swratioBest}{RGB}{241,181,34}
\definecolor{swratioBestEdge}{RGB}{163,105,0}
\definecolor{swratioEndpoint}{RGB}{95,95,95}
\begin{figure}[t]
\centering
  \begin{tikzpicture}
    \begin{axis}[
      width=0.94\linewidth,
      height=0.52\linewidth,
      xmin=-0.03,
      xmax=1.03,
      ymin=5.15,
      ymax=5.68,
      xtick={0,0.1,0.2,0.3,0.4,0.5,0.6,0.7,0.8,0.9,1},
      xticklabels={0,.1,.2,.3,.4,.5,.6,.7,.8,.9,1},
      xlabel={SW ratio $\rho$},
      ylabel={AL},
      grid=major,
      grid style={black!12},
      axis line style={black!65},
      tick align=outside,
      label style={font=\footnotesize},
      tick label style={font=\scriptsize},
      every axis plot/.append style={line width=1.15pt,mark size=2.5pt},
    ]
      \addplot[color=swratioEndpoint,dashed,forget plot] coordinates {
        (1,5.15) (1,5.68)
      };
      \addplot[color=swratioLine,mark=none] coordinates {
        (0.10,5.2748) (0.20,5.4814) (0.30,5.5541)
        (0.40,5.6044) (0.50,5.6141) (0.60,5.5766)
        (0.70,5.5374) (0.80,5.5329) (0.90,5.5096)
        (0.95,5.4656) (1,5.2300)
      };
      \addplot[only marks,color=swratioLine,mark=*] coordinates {
        (0.10,5.2748) (0.20,5.4814) (0.30,5.5541)
        (0.40,5.6044) (0.60,5.5766) (0.70,5.5374)
        (0.80,5.5329) (0.90,5.5096) (0.95,5.4656)
        (1,5.2300)
      };
      \node[circle,fill=white,minimum size=11pt,inner sep=0pt]
        at (axis cs:0.50,5.6141) {};
      \node[star,star points=5,star point ratio=1.8,
        fill=swratioBest,draw=swratioBestEdge,line width=0.45pt,
        minimum size=10pt,inner sep=0pt]
        at (axis cs:0.50,5.6141) {};
      \node[anchor=north west,font=\scriptsize]
        at (axis cs:0.025,5.665) {$\rho=0.5$: $5.614$};
      \draw[color=swratioEndpoint,dashed,line width=1pt]
        (axis cs:0.10,5.2748) -- (axis cs:0.015,5.150);
      \node[anchor=south west,font=\scriptsize,
        fill=white,fill opacity=0.85,text opacity=1,inner sep=1pt]
        at (axis cs:0.012,5.155) {SnapKV (4.073)};
      \node[anchor=south east,font=\scriptsize]
        at (axis cs:0.985,5.230) {SWA (5.230)};
    \end{axis}
  \end{tikzpicture}
\caption{Effect of the sliding-window allocation ratio. $4K$ OSL and $8K$ KV budget. Qwen3.5-4B drafts Qwen3.5-122B on SPEED-Bench 32K ISL data. $\rho=0$ denotes SnapKV with a fixed 64-token SW, and $\rho=1$ corresponds to pure SWA.}
\label{fig:effect-sw-ratio}
\end{figure}
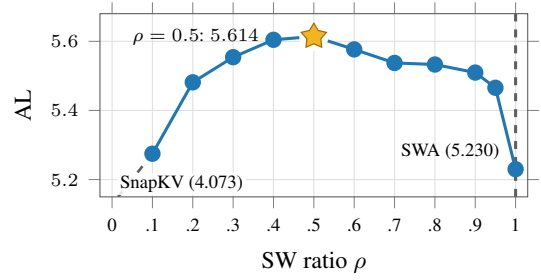

\textbf{Block-Level \riftstat.}
Serving systems store KVs in pages~\cite{kwon2023efficientmemorymanagementlarge}. We therefore divide the cache into contiguous blocks, each scored by the mean of its keys' scores. For an analysis of the block size effect on drafting performance, we refer the reader to Appendix~\ref{app:block-size-ablation}.

\subsection{\riftdyn: KV Cache Refresh During Decode}
\label{sec:dynamic-snapkv}

While \riftstat introduces gains in the draft AL, its chosen prefill KVs are still selected only once before decode. However, as generation progresses, the relevant prompt KVs dynamically change.
We therefore assume access to the full draft KV cache in HBM or offloaded memory for full KV rescoring.
We propose \riftdyn, which iteratively rescores the working KV cache every $\reviewfix{\Delta}$ decode steps. The resulting working KV cache is a combination of the decode-updated KV cache and the SW component of size $\lfloor \rho B\rfloor$. We focus on SnapKV-based scoring. However, \riftdyn is compatible with any alternative KV scoring functions.
\riftdyn refreshes the draft's working memory.
Its effectiveness stems from two main advantages: (i) it scores the entire available KV cache (excluding the SW component), and (ii) it computes scores using a recent observation window, so they better reflect the model's current state.

\riftdyn introduces a significant and consistent improvement in the draft's
\acceptancelength (see Section~\ref{sec:results} for results).  For example, with a refresh interval of $\reviewfix{\Delta}=128$ and a Qwen3.5-4B drafter at 100K ISL, \riftdyn retains $95.7\%$ of full-KV AL, compared with $85.6\%$ for \riftstat.
More frequent KV cache refreshes improve draft quality, with projected rescoring latency overhead  (As seen in Figure~\ref{fig:effect-dynamic-refresh}). We next consider adaptive refresh to improve the AL-refresh tradeoff.

\definecolor{refreshLine}{RGB}{31,119,180}
\definecolor{refreshBase}{RGB}{95,95,95}
\definecolor{refreshTime}{RGB}{230,126,34}

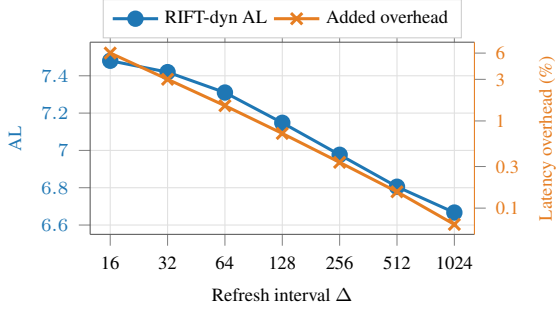
\begin{figure}[t]
  \centering
  \begin{tikzpicture}
    \begin{axis}[
      scale only axis,
      width=0.66\linewidth,
      height=0.33\linewidth,
      xmin=-0.35,
      xmax=6.35,
      ymin=6.55,
      ymax=7.58,
      xtick={0,1,2,3,4,5,6},
      xticklabels={16,32,64,128,256,512,1024},
      xlabel={Refresh interval $\Delta$},
      ylabel={AL},
      axis y line*=left,
      ylabel style={color=refreshLine},
      yticklabel style={color=refreshLine},
      grid=major,
      grid style={black!12},
      axis line style={black!65},
      tick align=outside,
      label style={font=\scriptsize},
      tick label style={font=\scriptsize},
        legend style={draw=black!20,fill=white,font=\scriptsize,
        at={(0.5,1.02)},anchor=south,legend columns=2},
      every axis plot/.append style={line width=1.15pt,mark size=2.5pt},
    ]
      \addplot[color=refreshLine,mark=*] coordinates {
        (0,7.4806) (1,7.4196) (2,7.3102) (3,7.1480)
        (4,6.9766) (5,6.8053) (6,6.6669)
      };
      \addlegendentry{\riftdyn AL}
      \addlegendimage{color=refreshTime,line width=1.15pt,mark=x,mark size=3pt}
      \addlegendentry{Added overhead}
    \end{axis}
    \begin{axis}[
      scale only axis,
      width=0.66\linewidth,
      height=0.33\linewidth,
      xmin=-0.35, xmax=6.35,
      ymode=log,
      ymin=0.05, ymax=8,
      axis x line=none,
      axis y line*=right,
      ylabel={Latency overhead (\%)},
      ylabel style={color=refreshTime},
      yticklabel style={color=refreshTime},
      ytick={0.1,0.3,1,3,6},
      yticklabels={0.1,0.3,1,3,6},
      tick align=outside,
      axis line style={refreshTime},
      label style={font=\scriptsize},
      tick label style={font=\scriptsize},
    ]
      \addplot[color=refreshTime,line width=1.15pt,mark=x, mark size=3pt] coordinates {
        (0,6.012) (1,3.004) (2,1.496) (3,0.719) (4,0.335) (5,0.154) (6,0.065)
      };
    \end{axis}
  \end{tikzpicture}
  \caption{Effect of \riftdyn's refresh interval on AL (blue) and projected rescoring overhead relative to measured draft decode time (orange). Qwen3.5-4B, 100K ISL, 8K cache budget, and 4K OSL. Appendix~\ref{app:rescoring-overhead} describes the latency projection.}

  \label{fig:effect-dynamic-refresh}
\end{figure}

\subsection{Adaptive KV cache Refresh}
\label{sec:adaptive-snapkv}

\riftdyn improves draft accuracy by periodically refreshing the KV cache, even when the current selection remains useful.
To avoid unnecessary refreshes, we propose \riftada, which adaptively chooses when to refresh the KV cache based on signals collected at the draft during decode.

At decode step $t$, let $\tau$ be the decode position of the most recent KV scoring and $a_t=t-\tau$ the age of the current cache selection. Then, a KV rescore occurs if a threshold policy $\reviewfix{\pi(z_t)}$ is activated or $a_t$ exceeds a maximal no-refresh interval $a_{\max}$:
$$
 u_t=\Ind\big\{(a_t\geq a_{\max})\lor\big(\pi(z_t) \land a_t\geq a_{\min}\big)\big\},
$$
where $u_t=1$ triggers a rescore. The minimum age $a_{\min}$ prevents consecutive refreshes, while the maximum age $a_{\max}$ bounds staleness. We evaluate $u_t$ every $d=8$ steps.

\textbf{Threshold Policies.}
At each $d$-token decision interval, we aggregate several signals $z_t$: an exponentially weighted mean of previous measured acceptance lengths $A_t$, draft entropy $H_t$, and query drift, which is given by $D^q_t = 1-\bar q_t\cdot \bar q_{\tau}$ and $\bar q$ denotes a normalized query summary. We average $D^q_t$ over full-attention layers and KV-head groups.
We compare each of these signals with a calibrated threshold (See Appendix~\ref{app:adaptive-refresh-audit} for details on the calibration process), and aggregate multiple scores into a threshold policy using (\texttt{and}/\texttt{or}) gates or a fitted weighted score. 
Figure~\ref{fig:adaptive-refresh-monotonic-74} illustrates how \riftada varies refresh frequency during generation, compared with \riftdyn's periodic schedule, showing how various stages in generation require different rescoring cadence.
The evaluated policies are detailed in Section~\ref{sec:results}.

\definecolor{tracePeriodic}{RGB}{92,102,112}
\definecolor{traceAdaptive}{RGB}{0,114,178}
\definecolor{traceScore}{RGB}{213,94,0}

\pgfplotsset{
  adaptive refresh trace/.style={
    width=0.8\linewidth,
    xmin=0,
    xmax=4096,
    xtick={0,1024,2048,3072,4096},
    axis line style={black!65},
    tick align=outside,
    tick label style={font=\scriptsize},
    label style={font=\scriptsize},
    ylabel style={font=\scriptsize,rotate=-90,anchor=east,xshift=-0.35em},
    xmajorgrids,
    grid style={black!10},
    enlarge x limits=false,
  },
  refresh heartbeat/.style={
    ycomb,
    line width=1.15pt,
    mark=*,
    mark size=1.7pt,
  },
  refresh score/.style={
    color=traceScore,
    line width=1.15pt,
    no marks,
  },
}

\begin{figure*}[!tbp]
  \centering
  \begin{tikzpicture}
    \begin{groupplot}[
      adaptive refresh trace,
      group style={group size=1 by 3,vertical sep=0.22cm},
    ]
      \nextgroupplot[
        height=0.12\linewidth,
        ymin=0,
        ymax=1.08,
        ytick=\empty,
        xticklabel=\empty,
        ylabel={Periodic (\riftdyn)},
      ]
        \addplot+[refresh heartbeat,color=tracePeriodic]
          table[x=position,y=value]
          {assets/data/adaptive-refresh-monotonic-74-periodic.dat};

      \nextgroupplot[
        height=0.12\linewidth,
        ymin=0,
        ymax=1.08,
        ytick=\empty,
        xticklabel=\empty,
        ylabel={Adaptive (\riftada)},
      ]
        \addplot+[refresh heartbeat,color=traceAdaptive]
          table[x=position,y=value]
          {assets/data/adaptive-refresh-monotonic-74-adaptive.dat};

      \nextgroupplot[
        height=0.15\linewidth,
        ymin=-6.4,
        ymax=1.0,
        ytick=\empty,
        xticklabels={0,1K,2K,3K,4K},
        xlabel={Decode position},
        ylabel={Score},
        ymajorgrids,
      ]
        \addplot[refresh score]
          table[x=position,y=score]
          {assets/data/adaptive-refresh-monotonic-74-score.dat};
        \addplot[black!65,dashed,line width=0.9pt]
          coordinates {(0,0) (4096,0)};
        \addplot[only marks,mark=*,mark size=1.5pt,color=black]
          table[x=position,y=score]
          {assets/data/adaptive-refresh-monotonic-74-refresh-score.dat};
    \end{groupplot}
  \end{tikzpicture}
  \caption{KV rescore patterns comparison on a candidate sample from SPEED-Bench. \riftada attains a higher AL and requires less KV refreshes. It can be seen that KV refreshes are not periodic and depend on generation location.
  }
  \label{fig:adaptive-refresh-monotonic-74}
\end{figure*}
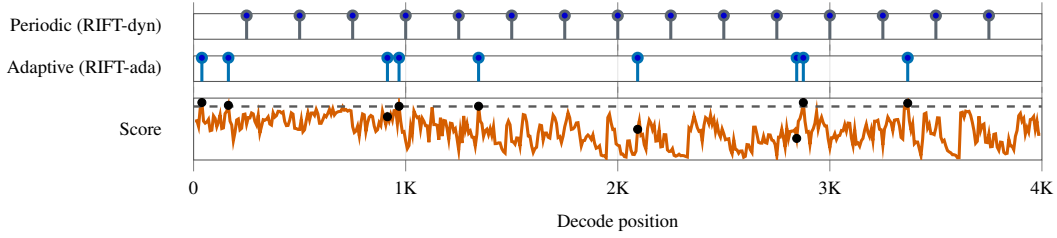

\subsection{Results}
\label{sec:results}
\colorlet{sectionfourrevision}{black}
\newcommand{\sectionfouredit}[1]{#1}
We evaluate \rift on SPEED-Bench, which comprises 8 different generation benchmarks, focusing on its throughput splits. Unless stated otherwise, we use draft length of $11$, maximum OSL of $4$K and decode temperatures $T\in\{0,1\}$. The KV cache budget is $8$K and accuracy is measured in average \acceptancelength. Additional results are provided in Appendix~\ref{app:section4-results}, where we cover more drafter-target pairs, datasets and generation setups. All \rift variations use $\rho=0.5$. 
For \riftada, we use the same refresh framework throughout and consider a threshold that results in strictly fewer refreshes than \riftdyn (see Appendix~\ref{app:adaptive-refresh-audit} for more details).
We also provide preliminary results on \rift applied beyond speculative decoding in Appendix~\ref{app:long-context-quality-protocol}: we evaluate \riftstat on a long-input, long-output reasoning benchmark, showing that its advantage extends beyond speculative decoding.

\begin{table}[tbp]
  \centering
  \footnotesize
  \setlength{\tabcolsep}{2pt}

  \begin{tabular}{llccccc}
    \toprule
    Draft (ISL) & $T$ & SW & SnapKV & \multicolumn{3}{c}{RIFT} \\\cmidrule(lr){5-7}
& & & & stat & dyn & ada \\
    \midrule
    \multirow{2}{*}{Qwen3.5 (32K)}
      & 0 & 5.23 & 4.07 & 5.62 & 6.12 & 6.19 \\
      & 1 & 4.70 & 3.69 & 4.99 & 5.48 & 5.55 \\
    \midrule
    Qwen3.5 (100K) & 0 & 6.12 & 4.46 & 6.40 & 7.14 & 7.17 \\
    Qwen3.5 (100K) & 1 & 5.40 & 3.82 & 5.55 & 6.21 & 6.36 \\
    \midrule
    \multirow{2}{*}{Llama3.1 (32K)}
      & 0 & 2.24 & 5.19 & 6.07 & 6.91 & $ 7.48$ \\
      & 1 & 1.55 & 3.83 & 4.54 & 5.05 & $ 5.52$ \\
    \midrule
    \multirow{2}{*}{N3.5 (self) (32K)}
      & 0 & 5.73 & 4.83 & 6.72 & 8.26 & 8.71 \\
      & 1 & 5.56 & 4.51 & 6.84 & 8.70 & 8.93 \\
    \midrule
    \multirow{2}{*}{\shortstack[l]{N3.5 (ext.) (32K)}}
      & 0 & 4.31 &  3.64 & 4.35 & 4.99 & 4.98 \\
      & 1 & 3.86 &  3.22 & 3.93 & 4.45 & 4.99 \\
    \bottomrule
  \end{tabular}
  \hypersetup{linkcolor=introrevision}
  \caption{AL on low-entropy SPEED-Bench 32K/100K ISL, with an 8K budget, 4K OSL, and draft length 11. Qwen3.5: 4B $\rightarrow$ 122B; Llama3.1: 8B $\rightarrow$ 70B; Nemotron3.5-Lightning in self-speculation and as an external draft to Nemotron3-Ultra target.}
  \label{tab:section4-model-temperature-relative}
\end{table}

\textbf{AL vs. Budget.} We start by demonstrating the accuracy of the various \rift variants across several evaluation settings.
First, we consider Qwen3.5-4B drafting for Qwen3.5-122B. Figure~\ref{fig:qwen35-overview}(b) compares \rift with SnapKV and SWA on SPEED-Bench 32K ISL across cache budgets. Figure~\ref{fig:qwen35-overview}(a) depicts AL vs. decode position, showing \riftstat benefits from both worlds: it preserves SnapKV's high early-generation AL while benefiting from the SW component later, as expected from Section~\ref{sec:root-cause}. Table~\ref{tab:section4-model-temperature-relative} extends the comparison to 100K ISL and additional model pairs. Specifically, we consider Nemotron3.5-Lightning (self-speculation and drafting for Nemotron3-Ultra) and Llama3.1-8B (Llama3.1-70B target), which is a full-attention model. For Qwen3.5 at 100K ISL, an 8K budget, and $T=0$, \riftstat, \riftdyn, and \riftada retain $85.6\%$, $95.5\%$, and $96.0\%$, respectively, compared with $59.7\%$ for SnapKV. Appendix~\ref{app:section4-results} provides additional results, encompassing various budgets, models, OSLs and workloads. Appendix~\ref{app:domains-breakdown} demonstrates \rift's advantage across various generation tasks.

\textbf{Beyond attention scores.}
While we demonstrate \riftstat on top of SnapKV, it can be used with any other prefill-based KV compression method. We combine several additional KV selection scores with \riftstat:
1) \textit{CriticalKV}'s output-perturbation score~\cite{feng2026criticalkv}, 2) \textit{Expected Attention}'s distribution over future queries~\cite{devoto2025expectedattention}, and 3) \textit{CurDKV}'s leverage scores~\cite{sengupta2025curdkv}.
We measure the benefit of applying \riftstat on top of each method. As seen in Table~\ref{tab:static-selectors-sw-ratio}, all three methods improve substantially at both budgets (see Table~\ref{tab:static-selectors-sw-ratio-full} for $T=1$).
The proposed observations naturally extend to direct variants of SnapKV, such as Ada-KV and PyramidKV, which focus on layer/head budget allocation~\cite{feng2025adakv,cai2024pyramidkv}.

\begin{table}[tbp]
\centering
\color{reviewfixcolor}
\captionsetup{labelfont={color=reviewfixcolor},textfont={color=reviewfixcolor}}
\footnotesize
\setlength{\tabcolsep}{3pt}
\begin{tabular}{llcc}
\toprule
Budget & Selection score & Baseline & + \riftstat \\
\midrule
4K & CriticalKV & 3.57 & 5.02 (+40.6\%) \\
 & Expected Attention & 3.59 & 5.03 (+40.2\%) \\
 & CurDKV & 3.29 & 4.84 (+47.1\%) \\
\midrule
8K & CriticalKV & 4.13 & 5.62 (+36.2\%) \\
 & Expected Attention & 4.06 & 5.56 (+36.9\%) \\
 & CurDKV & 3.76 & 5.35 (+42.2\%) \\
\bottomrule
\end{tabular}
\caption{Absolute AL at $T=0$ (relative gains in parentheses): fixed 64-token SW versus \riftstat ($\rho=0.5$). Qwen3.5-4B drafts Qwen3.5-122B on SPEED-Bench 32K ISL, 4K OSL, draft length 11.}
\label{tab:static-selectors-sw-ratio}
\end{table}

\textbf{Cross-Family Drafting}
We consider a cross-family external drafting setting. The main challenge in this setup is that the draft and target models do not share the same vocabulary. We adapt drafters through a simple tokenizer matching step followed by target-model KD (see Appendix~\ref{app:tokenizer-adaptation} for tokenizer-adaptation details). We consider several cross-family draft-target pairs and apply \rift in a long-ISL drafting setting. Specifically, we consider Llama3.2-1B and Qwen3.5-4B drafts and DeepSeek-V4 (DSV4) Pro and Flash as target models.
As seen in Table~\ref{tab:cross-family-main}, \rift extends to the cross-model scenario, supporting external drafting when an in-family smaller model is unavailable.

\begin{table}[tbp]
\centering
\footnotesize
\setlength{\tabcolsep}{1.0pt}
\begin{tabular}{llccc}
\toprule
Draft & Target & SnapKV & + \riftstat & + \riftdyn \\
\midrule
\multirow{2}{*}{Qwen3.5-4B} & DSV4-Flash & 3.50 & 4.26 & 5.35 \\
 & DSV4-Pro & 3.43 & 4.26 & 5.07 \\
\midrule
\multirow{2}{*}{Llama3.2-1B} & DSV4-Flash & 2.50 & 3.24 & 3.50 \\
 & DSV4-Pro & 2.37 & 3.07 & 3.26 \\
\bottomrule
\end{tabular}
\caption{Absolute AL for adapted drafters of DeepSeek-V4-Flash/Pro on low-entropy SPEED-Bench: 32K ISL, 8K budget, 4K OSL, $T=0$, draft length 11.}
\label{tab:cross-family-main}
\end{table}

\textbf{The benefit of adaptive refresh.}
We compare \riftada with \riftdyn's periodic refresh through their AL--refresh tradeoff, measuring AL gains at matched refresh counts and refresh reductions at matched AL.
The policy triggers a refresh if any of the following conditions holds: acceptance is at or below its threshold, entropy is at or above its threshold, acceptance has fallen or entropy has risen sufficiently relative to post-refresh values, or query drift exceeds its threshold.
As seen in Figure~\ref{fig:adaptive-snapkv-pareto}, this policy, which sweeps the query drift threshold, achieves $2.03\times$ fewer refreshes (or up to $+0.122$ AL at the same refresh count). We refer the reader to Appendix~\ref{app:adaptive-snapkv-results} for additional settings, parameter-transfer analysis and policy information.
\definecolor{adaptiveFixed}{RGB}{95,95,95}
\definecolor{adaptiveRobust}{RGB}{0,114,178}
\definecolor{adaptiveKnee}{RGB}{230,159,0}
\definecolor{adaptiveBest}{RGB}{204,51,17}

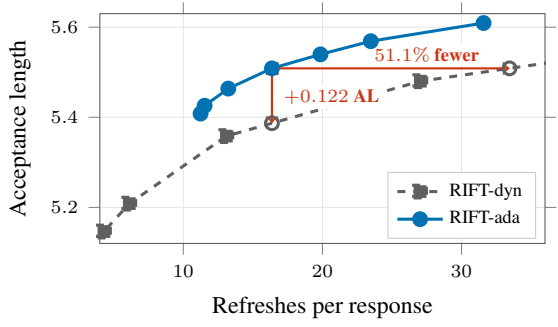
\begin{figure}[tbp]
\centering
\begin{tikzpicture}
      \begin{axis}[
      width=0.97\linewidth,
      height=0.6\linewidth,
      xlabel={Refreshes per response},
      grid=major,
      grid style={black!10},
      axis line style={black!65},
      tick align=outside,
      label style={font=\small},
      tick label style={font=\scriptsize},
      title style={font=\small\bfseries},
      legend style={draw=black!20,fill=white,font=\scriptsize,
        cells={anchor=west}},
      every axis plot/.append style={line width=1.15pt,mark size=2.4pt},
        ylabel={Acceptance length},
        xmin=4,xmax=36,
        ymin=5.12,ymax=5.63,
        legend style={at={(0.98,0.02)},anchor=south east},
      ]
        \addplot[color=adaptiveFixed,dashed,mark=square*] coordinates {
          (0,4.9923) (4.2969,5.1464) (6.1172,5.2083)
          (13.0938,5.3582) (27.0391,5.4801) (54.9453,5.6041)
        };
        \addlegendentry{\riftdyn}
        \addplot[color=adaptiveRobust,mark=*] coordinates {
          (11.2422,5.4081) (11.5313,5.4257) (13.2344,5.4637)
          (16.3711,5.5086) (19.8594,5.5396) (23.4766,5.5689)
          (31.5820,5.6092)
        };
        \addlegendentry{\riftada}
        \addplot[color=adaptiveFixed,only marks,mark=o]
          coordinates {(16.3711,5.3868) (33.4475,5.5086)};
        \draw[<->,adaptiveBest,line width=0.8pt]
          (axis cs:16.3711,5.3868) -- (axis cs:16.3711,5.5086);
        \draw[<->,adaptiveBest,line width=0.8pt]
          (axis cs:16.3711,5.5086) -- (axis cs:33.4475,5.5086);
        \node[anchor=west,font=\scriptsize\bfseries,color=adaptiveBest,
          fill=white,fill opacity=0.98,text opacity=1,inner sep=1.5pt]
          at (axis cs:16.9,5.445) {$+0.122$ AL};
        \node[anchor=south,font=\scriptsize\bfseries,color=adaptiveBest,
          fill=white,fill opacity=0.98,text opacity=1,inner sep=1.5pt]
          at (axis cs:27.5,5.512) {$51.1\%$ fewer};
\end{axis}
\end{tikzpicture}
\caption{\riftada AL-refresh tradeoff at $T=1$ for Qwen3.5-4B drafting Qwen3.5-122B on SPEED-Bench 32K low-entropy data, with an 8K budget and 4K OSL.}
\label{fig:adaptive-snapkv-pareto}
\end{figure}

\textbf{Comparison with additional baselines.}
We compare \rift with StreamingLLM \cite{xiao2024streamingllm}, H$_2$O~\cite{zhang2023h2o} and TriAttention~\cite{mao2026triattention}, which are decode-time KV selection methods.
As seen in Table~\ref{tab:decode-baselines-main}, StreamingLLM is outperformed by all compression methods, H$_2$O outperforms \riftstat, while \riftdyn and \riftada achieve higher AL than either baseline across the two context lengths and temperatures.
Appendix~\ref{app:decode-policy-comparison} provides comparison across additional settings.

\begin{table}[tbp]
\centering
\footnotesize
\setlength{\tabcolsep}{5pt}
\begin{tabular}{lcccc}
\toprule
 & \multicolumn{2}{c}{32K ISL} & \multicolumn{2}{c}{100K ISL} \\
\cmidrule(lr){2-3}\cmidrule(lr){4-5}
Method & $T=0$ & $T=1$ & $T=0$ & $T=1$ \\
\midrule
StreamingLLM & 5.34 & 4.73 & 5.60 & 4.98 \\
H$_2$O & 5.73 & 5.03 & 6.65 & 5.77 \\
TriAttention & 5.43 & 4.77 & 6.10 & 5.11 \\
\riftstat & 5.62 & 4.99 & 6.40 & 5.55 \\
\riftdyn & 6.12 & 5.48 & 7.14 & 6.21 \\
\riftada & \textbf{6.17} & \textbf{5.51} & \textbf{7.17} & \textbf{6.36} \\
\bottomrule
\end{tabular}
\caption{Absolute AL against decode-time baselines. Qwen3.5-4B draft, low-entropy SPEED-Bench, 8K budget, 4K OSL, and draft length 11.}
\label{tab:decode-baselines-main}
\end{table}

\section{Improving Draft KV Compressibility with Training}
\label{sec:draft-compressibility-training}
\begin{table*}[t]
\centering
\small
\setlength{\tabcolsep}{3pt}
\begin{tabular}{ll@{\hspace{4pt}}ccc@{\hspace{4pt}}ccc}
\toprule
& & \multicolumn{3}{c}{\riftstat}
  & \multicolumn{3}{c}{\riftdyn} \\
\cmidrule(lr){3-5}\cmidrule(lr){6-8}
Draft $\rightarrow$ target & $T$
& 2K & 4K & 8K & 2K & 4K & 8K \\
\midrule
\multirow{2}{*}{Qwen3.5 (4B $\rightarrow$ 122B)}
& 0 & $+7.70\%$ & $+6.87\%$ & $+5.70\%$
    & $+7.48\%$ & $+5.50\%$ & $+3.72\%$ \\
& 1 & $+16.19\%$ & $+14.88\%$ & $+12.28\%$
    & $+15.51\%$ & $+12.25\%$ & $+9.21\%$ \\
\midrule
\multirow{2}{*}{Lightning $\rightarrow$ Nemotron3-Ultra}
& 0 & $+7.51\%$ & $+7.48\%$ & $+7.59\%$
    & $+7.88\%$ & $+7.86\%$ & $+5.59\%$ \\
& 1 & $+15.69\%$ & $+15.63\%$ & $+15.08\%$
    & $+16.42\%$ & $+15.84\%$ & $+13.19\%$ \\
\midrule
\multirow{2}{*}{Lightning self-speculation}
& 0 & $+5.3\%$ & $+3.4\%$ & $+2.5\%$
    & $+6.5\%$ & $+2.8\%$ & $+1.0\%$ \\
& 1 & $+11.0\%$ & $+7.8\%$ & $+4.2\%$
    & $+10.3\%$ & $+5.0\%$ & $+1.5\%$ \\
\bottomrule
\end{tabular}
\caption{Relative AL improvement from compression-aware KD over the base model under the same KV compression policy and budget, evaluated on SPEED-Bench 32k ISL with draft length 11.
}
\label{tab:trainable-snapkv-improvement}
\end{table*}

\colorlet{drafttrainrevision}{naacleditcolor}

The previous section improves KV compression at inference with a fixed draft. A natural question is:

\begin{center}
\textit{Can we further improve the drafter's KV compressibility using dedicated training?}
\end{center}

Inspired by recent results in compressibility training of pretrained models~\cite{gelberg2026kvcat}, we propose a draft compressibility post-training phase.
Our goal is to minimize the gap between the draft's working KV cache in training and during inference.
In what follows, we focus on compressibility using attention scores. However, the training method itself is score-agnostic.
The resulting objective promotes target-aligned proposals under a restricted KV budget.

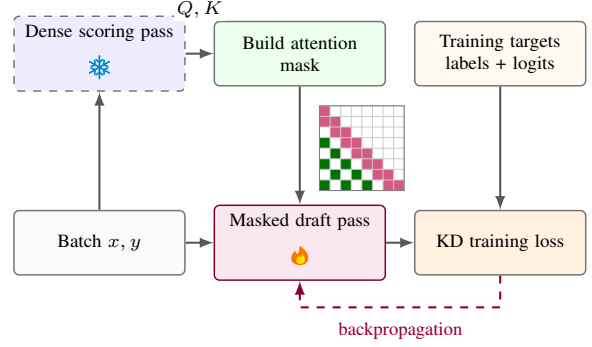
\begin{figure}[t]
\centering
\begin{tikzpicture}[
font=\scriptsize,
b/.style={draw=black!55,line width=.55pt,rounded corners=2pt,align=center,minimum height=.85cm,text width=2.05cm,inner sep=3pt},
a/.style={-latex,line width=.8pt,draw=black!65},
el/.style={font=\scriptsize,fill=white,inner sep=1pt}]
\node[b,fill=blue!7,dashed] (score) at (0,2.5) {Dense scoring pass\\[7pt]\strut};
\node[b,fill=green!7] (mask) at (2.65,2.5) {Build attention\\mask};
\node[b,fill=orange!5] (targets) at (5.3,2.5) {Training targets\\labels + logits};
\node[b,fill=black!2] (batch) at (0,0) {Batch $x,y$};
\node[b,fill=purple!9,draw=purple!65!black] (student) at (2.65,0) {Masked draft pass\\[7pt]\strut};
\node[b,fill=orange!12] (loss) at (5.3,0) {KD training loss};

\begin{scope}[shift={(0,2.32)},draw=cyan!65!blue,line width=.65pt,line cap=round]
  \foreach \angle in {0,60,120,180,240,300} {
    \begin{scope}[rotate=\angle]
      \draw (0,0)--(0,.15);
      \draw (-.045,.10)--(0,.065)--(.045,.10);
    \end{scope}
  }
\end{scope}
\begin{scope}[shift={(2.65,-.24)},scale=.24]
  \path[fill=orange!90!red] (0,-.23)
    .. controls (-.65,-.20) and (-.53,.35) .. (-.24,.57)
    .. controls (-.25,.31) and (-.02,.28) .. (-.01,.83)
    .. controls (.59,.47) and (.66,-.21) .. (0,-.23);
  \path[fill=yellow!85!orange] (0,-.17)
    .. controls (-.27,-.13) and (-.22,.13) .. (-.04,.32)
    .. controls (-.01,.11) and (.15,.18) .. (.12,.43)
    .. controls (.39,.07) and (.29,-.17) .. (0,-.17);
\end{scope}

\draw[a] (batch)--(score);
\draw[a] (batch)--(student);
\draw[a] (score)--node[el,above,yshift=12pt] {$Q,K$}(mask);
\draw[a] (mask.south)--(student.north);
\begin{scope}[shift={(2.91,.69)},x=.14cm,y=.14cm]
  \fill[white] (0,0) rectangle (8,8);
  \foreach \x/\y in {0/7,1/6,2/5,3/4,4/3,5/2,6/1,7/0,
                      0/6,1/5,2/4,3/3,4/2,5/1,6/0} {
    \fill[purple!65] (\x,\y) rectangle ++(1,1);
  }
  \foreach \x/\y in {0/4,1/3,0/2,2/2,1/1,3/1,0/0,2/0,4/0} {
    \fill[green!45!black] (\x,\y) rectangle ++(1,1);
  }
  \draw[step=1,black!20,line width=.2pt] (0,0) grid (8,8);
  \draw[black!50,line width=.4pt] (0,0) rectangle (8,8);
\end{scope}
\draw[a] (student)--(loss);
\draw[a] (targets)--(loss);
\draw[a,dashed,draw=purple!75!black] (loss.south)--++(0,-.42)-| (student.south);
\node[font=\scriptsize,text=purple!75!black,anchor=north] at (3.98,-.92) {backpropagation};
\end{tikzpicture}
\caption{Compression-aware distillation. The snowflake and fire symbols denote frozen and trainable passes, respectively. The mask combines recent positions (purple) with importance-selected context (green). }
\label{fig:trainable-snapkv-forward}
\end{figure}

\textbf{Compression-aware knowledge distillation.}
We propose a modified KD forward pass.
We begin with a dense scoring pass by applying a forward pass to the frozen draft model. We capture post-RoPE queries and keys in each attention layer. We score keys by summing their attention scores against a fixed observation window.
We define a compression ratio $p$ and a SW ratio $\rho$. During training, the effective cache budget is $B=\lfloor pL\rfloor$, of which $r=\lfloor\rho B\rfloor$ positions are reserved for a SW component, where $L$ is the sample's sequence length. We build an attention mask that combines the SWA with the highest-scoring prompt positions, obtained from the dense scoring pass.
We then perform an additional student forward pass, in which we apply the LoRA adapters and the built attention masks to get a \textit{sparse} forward pass.
We combine cross-entropy and KL distillation using precomputed teacher rollouts and logits:
$$
\mathcal{L}
= \lambda_1\operatorname{CE}
   (q_\theta,y)
 + \lambda_2 D_{\mathrm{KL}}
   (p\|q_\theta),
$$
where $y$ is the teacher-generated next token, and
$q_\theta$ and $p$
are the student and teacher distributions. The student uses the sparse attention mask. 
Figure~\ref{fig:trainable-snapkv-forward} provides an overview. 
Our method extends to self-speculation by toggling the adapters and using the corresponding selection method during inference.

\subsection{Results}

\paragraph{Setting.}
We train for $10$K steps using the \textit{Nemotron V2 post-training-dataset},\footnote{https://huggingface.co/datasets/nvidia/Nemotron-Post-Training-Dataset-v2} with batch size $64$ and SW ratio $\rho=0.25$. To increase budget robustness we randomly sample compression ratios for each sample. Evaluation uses SPEED-Bench 32K low-entropy data, $4$K OSL, and draft length $11$. Appendix~\ref{app:training-configuration} gives the full recipe.

\textbf{Compression-aware gains.}
We evaluate Qwen3.5 and Nemotron3.5-Lightning, considering both self speculation and external drafting.
As seen in Table~\ref{tab:trainable-snapkv-improvement}, compressibility training improves the drafter's performance under \rift. For example, we see gains of up to $16.19\%$ for Qwen3.5-4B (\riftstat, 2K budget, $T=1$) and $16.42\%$ gains for Nemotron3.5-Lightning (\riftdyn at 2K and $T=1$). Both comparisons measure improvement over the untrained drafter under the same compression policy and budget.
We note that gains are smaller at larger budgets, as the baseline method often already retains more than $95\%$ of the full-KV AL, leaving a very small gap for improvement.

\textbf{Benefit over no-mask training.}
To isolate the benefit of compression-aware training from that of the post-training data, we compare with the same pipeline without attention masking.
As seen in Table~\ref{tab:trainable-snapkv-no-mask-improvement}, compressibility training offers a significant improvement on top of the baseline training scheme, especially in the lower budget regimes.

\begin{table}[tbp]
\centering
\footnotesize
\setlength{\tabcolsep}{5pt}
\begin{tabular}{llccc}
\toprule
Method & $T$ & 2K & 4K & 8K \\
\midrule
\multirow{2}{*}{\riftstat}
& 0 & $+5.44\%$ & $+4.68\%$ & $+3.71\%$ \\
& 1 & $+7.31\%$ & $+6.37\%$ & $+4.66\%$ \\
\midrule
\multirow{2}{*}{\riftdyn}
& 0 & $+5.34\%$ & $+3.51\%$ & $+1.82\%$ \\
& 1 & $+7.25\%$ & $+4.73\%$ & $+2.37\%$ \\
\bottomrule
\end{tabular}
  \caption{Relative \acceptancelength improvement from compression-aware training over a similar training recipe with no
  compression-driven masks. Qwen3.5-4B$\to$Qwen3.5-122B, SPEED-Bench 32k ISL, draft length 11. 
  }
  \label{tab:trainable-snapkv-no-mask-improvement}
\end{table}

\section{Related Work}
\label{sec:related-work}
Due to the amount of existing methods, we describe the most closely related work to our methods. We provide a broader discussion of literature in Appendix~\ref{app:related-work}. SnapKV estimates prompt importance from observation-window attention and PyramidKV and AdaKV refine allocation across layers and heads~\cite{li2024snapkv,cai2024pyramidkv,feng2025adakv}. These methods optimize \emph{which} KVs survive compression, while we study how prefill decisions become stale during long generation. Decode-time policies continually adapt the working cache, e.g., StreamingLLM retains attention sinks and recent tokens, while H$_2$O combines recent tokens with accumulated-attention heavy hitters~\cite{xiao2024streamingllm,zhang2023h2o}.
TriForce uses a retrieved target cache as an intermediate drafter, refreshed periodically or by acceptance gating~\cite{sun2024triforce}. MagicDec studies sparse-KV self-speculation at high concurrency. SpecExtend, SparseSpec, and Vegas study how verification can be used for KV cache compression~\cite{sadhukhan2025magicdec,cha2025specextend,zhao2025sparsespec,yue2026vegas}. In compressibility training, KV-CAT~\cite{gelberg2026kvcat} proposes a continual pretraining phase with a learned KV scoring model.

\section{Conclusion}
\label{sec:conclusion}
In this work, we identified decode staleness, where prefill-selected context loses relevance as OSL grows, especially in hybrid attention architectures. We show that decode staleness degrades the draft-target alignment in speculative decoding. We proposed \rift: \riftstat combines recency with important prompt positions, \riftdyn improves KV selection by rescoring the cache during decode and \riftada allows for adaptive KV rescore to minimize the induced latency overhead. To  improve \rift we propose a compression-aware training phase. Together, our results demonstrate the value of adapting both KV cache selection policies and draft models to retain acceptance lengths under fixed KV cache budgets.
A promising future direction is to combine \rift's draft-side staleness signals with target-verification attention, as initial signals show promising results in this direction~\cite[e.g.,][]{cha2025specextend,yue2026vegas}.

\FloatBarrier
\section*{Limitations}
At the time of writing, external draft models generally provide lower end-to-end speedups in GPU-only deployments than tail-augmented drafters such as MTP, EAGLE, and DFlash. Our work is therefore motivated by future heterogeneous systems in which drafting runs on a memory-constrained, SRAM-based accelerator, an emerging direction for which \cite{aubrey2026groq3lpx} provides an early signal. In such a system, KV-selection scores could be computed on the GPU using a copy of the draft model, while only the selected KV blocks are transferred to the drafting device. Our evaluation focuses on acceptance-length trade-offs under a fixed active draft KV cache budget. Because decode-time refresh assumes that the full draft KV history remains available in GPU HBM or offloaded memory, this active budget constrains the drafting device's memory footprint but not the system's total stored KV memory.
Furthermore, on a GPU setup with naive measurements, we see a relative cost overhead of up to $6.0\%$ of KV cache rescoring compared to the total draft decode latency. This overhead may become more significant when drafting is accelerated on an SRAM-based device, although its magnitude in disaggregated deployments remains to be measured.

\bibliography{references}

\clearpage
\appendix
\section*{Appendix for ``\naaclPaperTitle''}

\suppressfloats[t]
\makeatletter
\setlength{\@fptop}{0pt}
\setlength{\@fpsep}{12pt}
\setlength{\@fpbot}{0pt plus 1fil}
\setlength{\@dblfptop}{0pt}
\setlength{\@dblfpsep}{12pt}
\setlength{\@dblfpbot}{0pt plus 1fil}
\makeatother
\section{KV cache Costs in Hybrid Draft Models}
\label{app:draft-kv-memory}
\begin{figure*}[!bp]
    \centering
    \input{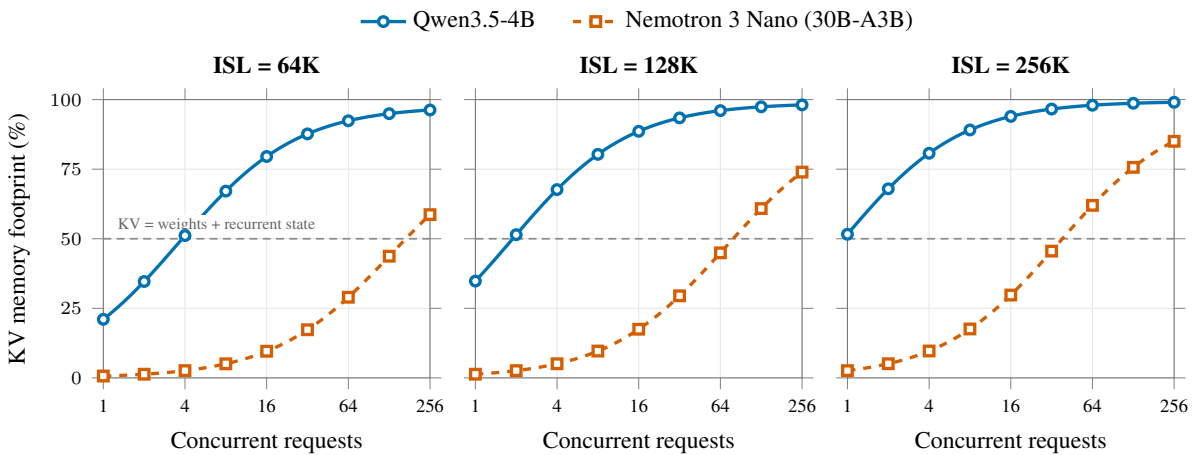}
    \caption{Attention KV cache share of total memory in two hybrid architectures, evaluated at the end of prefill.}
    \label{fig:draft-kv-memory}
\end{figure*}

We quantify the KV cache footprint in Qwen3.5-4B and
Nemotron-3-Nano (30B-A3B) relative to the combined memory (of weights, KV cache and recurrent states). Concurrent requests share the same weights but retain separate caches. KV memory grows with context length and concurrency while recurrent states (such as SSM and GDN states) grow with concurrency but not with ISL or OSL. We assume \texttt{bf16} weights and KV entries, \texttt{fp32} recurrent matrices, and \texttt{bf16}
convolution buffers, giving approximately 49.5\,MiB of state per request for Qwen and 47.1\,MiB for Nemotron3-Nano.
Figure~\ref{fig:draft-kv-memory} shows that long-context KV storage
can still dominate. For Qwen3.5-4B, four requests at 64K tokens use 51.1\% of the occupied memory, despite only 25\% of layers using full attention. Nano's smaller per-token cache and larger resident weight footprint delay the crossover. Nonetheless, at 256K ISL and 64 concurrent requests, its KV cache accounts for 62.0\% of the accounted memory. Consequently, it is clear that the KV memory bottleneck remains evident in standard serving loads.

\section{KV Cache Compression in Tail-Augmented Drafters}
\label{app:tail-augmented-kv}
We provide additional information on our initial results for KV compression's effect on MTP. Specifically, we examined Nemotron3-Super's MTP draft, which went through a post-training phase with a 1K SW. The training used the Nemotron post training V2 dataset, using a standard KD scheme for 10K training steps. The resulting MTP was evaluated on SPEED-Bench 32k and 100k ISL data.
The MTP's ALs were 4.24 (94\% of full KV AL) at 32K ISL and 5.07 (95\% of full KV AL) at 100K ISL. Consequently, a simple SWA shows sufficient AL retention in the settings of interest, rendering it a sufficient and simpler option over \rift for MTPs.
Furthermore, the overall reduction in the system's TPS due to the MTP's KV compression was negligible, implying that in a tail-augmented drafting setting, the MTP's KV cache may not pose as a significant bottleneck compared to external drafting and self-speculation.

\section{Additional Staleness Analysis}
\label{app:root-cause}
We provide additional evidence on draft staleness.
Figure~\ref{fig:rca-cross-architecture-position} extends the comparison across model families: SW outperforms SnapKV late in generation for GPT-OSS and Nemotron, as for Qwen3.5 in Figure~\ref{fig:qwen35-overview}(a), while Llama-3.1 consistently favors SnapKV over SW. Qwen3, which is a full-attention model, shows ambiguous results in terms of one method's dominance over the other.
Next, we evaluate the effect of reasoning. Figure~\ref{fig:rca-qwen-thinking-position} compares Qwen3.5-4B's drafting accuracy with thinking enabled and disabled (at both the draft and target models). Disabling thinking reduces the staleness gap in the final position bucket from $2.11$ to $0.60$ AL, although the SnapKV-SW crossover persists. All results use $T=0$, low-entropy SPEED-Bench 32k ISL data with a fixed 8K KV cache budget.

\begin{figure}[t]
\centering
\begin{tikzpicture}
\begin{axis}[width=.94\linewidth,height=.55\linewidth,xmin=0,xmax=5,ymin=0,ymax=85,
xtick={0,1,2,3,4,5},xticklabels={64,512,1K,2K,3K,4K},xlabel={Generated-token position},ylabel={Attention mass (\%)},
label style={font=\footnotesize},tick label style={font=\scriptsize},grid=major,grid style={black!12},
legend style={font=\scriptsize,legend columns=2,at={(.98,.98)},anchor=north east}]
        \addplot[color=rcaprompt,mark=square*] coordinates {
          (0,77.8389) (1,66.4315) (2,58.1513)
          (3,43.9660) (4,41.6979) (5,34.9848)
        };
        \addlegendentry{Prompt}
        \addplot[color=rcagenerated,mark=triangle*] coordinates {
          (0,22.1614) (1,33.5687) (2,41.8466)
          (3,56.0329) (4,58.3021) (5,65.0156)
        };
        \addlegendentry{Decode}

\end{axis}
\end{tikzpicture}
\caption{Full-attention mass shifts from prompt tokens to the generated prefix in the Qwen3.5-4B draft on 32K-ISL data.}
\label{fig:rca-prompt-decode}
\end{figure}
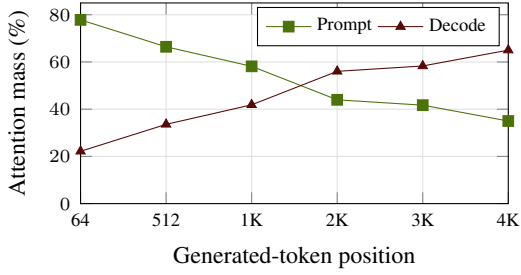
\definecolor{rcanative}{RGB}{55,55,55}
\definecolor{rcasw}{RGB}{31,119,180}
\definecolor{rcasnap}{RGB}{214,79,55}
\begin{figure*}[!tbp]
  \centering
  \begin{tikzpicture}
    \begin{groupplot}[
      group style={group size=4 by 1,horizontal sep=0.90cm},
      scale only axis,
      width=0.185\linewidth,
      height=0.17\linewidth,
      xmin=0,
      xmax=6,
      ymin=1.0,
      ymax=7.2,
      xtick={0,3,6},
      xticklabels={0--64,0.5--1k,3k+},
      x tick label style={rotate=30,anchor=east,font=\scriptsize},
      grid=major,
      grid style={black!12},
      tick align=outside,
      axis line style={black!65},
      every axis plot/.append style={line width=0.9pt,mark size=1.8pt},
      label style={font=\scriptsize},
      ylabel style={font=\scriptsize,at={(axis description cs:-0.20,0.5)}},
      y tick label style={font=\scriptsize},
      ylabel style={font=\scriptsize,at={(axis description cs:-0.20,0.5)}},
      title style={font=\scriptsize},
    ]
      \nextgroupplot[title={GPT-OSS},legend to name=rca-condensed-legend,legend columns=3,legend style={draw=black!20,font=\scriptsize},ylabel={AL},ymin=1.5,ymax=9.2]
        \addplot[color=rcanative,mark=*,forget plot] coordinates {
          (0,3.771976) (1,4.204034) (2,4.608589) (3,5.277034) (4,6.481914) (5,7.847751) (6,8.641463)
        };
        \addplot[color=rcasw,mark=square*,forget plot] coordinates {
          (0,2.717938) (1,3.392442) (2,3.818781) (3,4.440259)
          (4,4.886226) (5,4.584434) (6,4.088112)
        };
        \addplot[color=rcasnap,mark=triangle*,forget plot] coordinates {
          (0,3.645065) (1,3.158649) (2,2.817484) (3,2.529430)
          (4,1.935845) (5,1.761336) (6,1.702278)
        };

      \addlegendimage{color=rcanative,mark=*}
\addlegendentry{Full KV cache}
\addlegendimage{color=rcasw,mark=square*}
\addlegendentry{SW}
\addlegendimage{color=rcasnap,mark=triangle*}
\addlegendentry{SnapKV}
\nextgroupplot[title={Nemotron},ylabel={AL},ymin=3,ymax=8.4]
        \addplot[color=rcanative,mark=*] coordinates {
          (0,4.709316) (1,6.578810) (2,7.274135) (3,7.048083)
          (4,7.392578) (5,7.754072) (6,7.888351)
        };
\addplot[color=rcasw,mark=square*] coordinates {
          (0,4.546903) (1,5.704082) (2,5.863279) (3,5.404093)
          (4,4.995806) (5,5.047081) (6,4.995700)
        };
        \addplot[color=rcasnap,mark=triangle*] coordinates {
          (0,4.489081) (1,5.070813) (2,4.820939) (3,4.209988)
          (4,3.763728) (5,3.630381) (6,3.535347)
        };

      \nextgroupplot[title={Qwen3},ylabel={AL},ymin=2.3,ymax=8.2]
        \addplot[color=rcanative,mark=*] coordinates {
          (0,5.048934) (1,3.673398) (2,3.902572) (3,4.851249) (4,5.843494) (5,6.680074) (6,7.632488)
        };
        \addplot[color=rcasw,mark=square*] coordinates {
          (0,3.028986) (1,2.577777) (2,2.920566) (3,3.478226)
          (4,4.089363) (5,4.529083) (6,4.843076)
        };
        \addplot[color=rcasnap,mark=triangle*] coordinates {
          (0,4.949129) (1,3.107146) (2,2.868119) (3,3.158908)
          (4,3.504772) (5,3.604310) (6,3.917780)
        };

      \nextgroupplot[title={Llama-3.1},ylabel={AL},ymin=1,ymax=12]
        \addplot[color=rcanative,mark=*] coordinates {
          (0,6.337071) (1,10.109257) (2,10.162159) (3,10.236176)
          (4,10.392035) (5,11.068598) (6,11.149395)
        };
\addplot[color=rcasw,mark=square*] coordinates {
          (0,1.301039) (1,2.008697) (2,2.452141) (3,2.826169)
          (4,3.363094) (5,3.843003) (6,4.143074)
        };
        \addplot[color=rcasnap,mark=triangle*] coordinates {
          (0,5.324228) (1,6.680943) (2,6.078876) (3,6.127432)
          (4,6.017333) (5,6.146534) (6,6.094252)
        };

    \end{groupplot}
    \node[font=\scriptsize,anchor=north] at ($(group c1r1.south)!0.5!(group c4r1.south)+(0,-0.75cm)$) {Generated-token position};
    \node[anchor=south] at ($(group c1r1.north)!0.5!(group c4r1.north)+(0,0.5cm)$) {\pgfplotslegendfromname{rca-condensed-legend}};
  \end{tikzpicture}
  \caption{Interval AL across model families, with full-KV references. Target/draft pairs (left to right): GPT-OSS-120B/20B, Nemotron-Super-120B/Nano-30B, Qwen3-235B/8B, and Llama-3.1-70B/8B. Each panel uses its own AL scale. The corresponding Qwen3.5 comparison appears in Figure~\ref{fig:qwen35-overview}(a).}
  \label{fig:rca-cross-architecture-position}
  \label{fig:rca-hybrid-pairs-position}
\end{figure*}
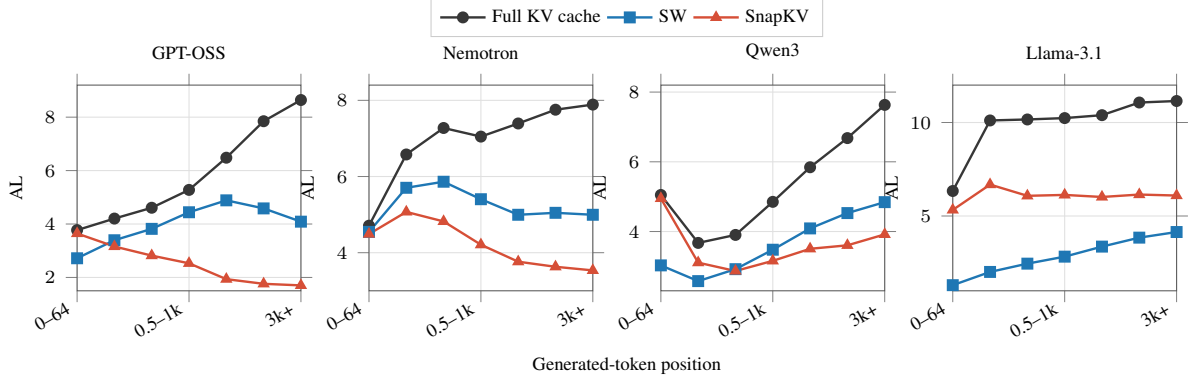

\definecolor{rcanative}{RGB}{55,55,55}
\definecolor{rcasw}{RGB}{31,119,180}
\definecolor{rcasnap}{RGB}{214,79,55}

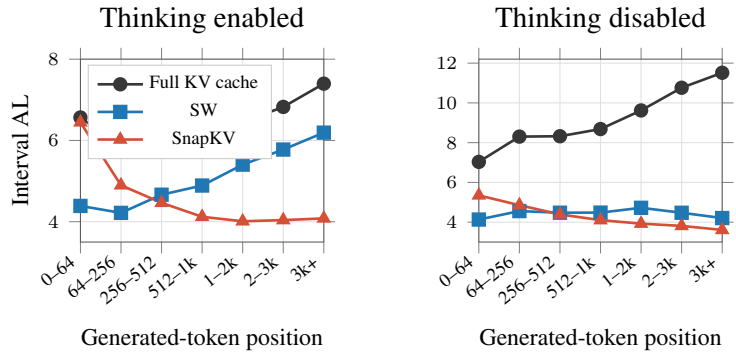
\begin{figure*}[!tbp]
  \centering
  \begin{tikzpicture}
    \begin{groupplot}[
      group style={group size=2 by 1,horizontal sep=2.05cm},
      width=0.3\linewidth,
      height=0.25\linewidth,
      xmin=0,
      xmax=6,
      xtick={0,1,2,3,4,5,6},
      xticklabels={0--64,64--256,256--512,512--1k,1--2k,2--3k,3k+},
      x tick label style={rotate=38,anchor=east,font=\scriptsize},
      xlabel={Generated-token position},
      grid=major,
      grid style={black!12},
      tick align=outside,
      axis line style={black!65},
      every axis plot/.append style={line width=1pt,mark size=2.2pt},
      label style={font=\small},
      y tick label style={font=\scriptsize},
    ]
      \nextgroupplot[
        title={Thinking enabled},
        ylabel={Interval AL},
        ymin=3.5,
        ymax=8.0,
        legend style={draw=black!20,fill=white,font=\scriptsize,at={(0.03,0.97)},anchor=north west},
      ]
        \addplot[color=rcanative,mark=*] coordinates {
          (0,6.564879) (1,5.844062) (2,6.077655) (3,6.017546)
          (4,6.447035) (5,6.824815) (6,7.396992)
        };
        \addlegendentry{Full KV cache}
        \addplot[color=rcasw,mark=square*] coordinates {
          (0,4.389059) (1,4.217744) (2,4.664816) (3,4.890495)
          (4,5.404534) (5,5.775898) (6,6.191576)
        };
        \addlegendentry{SW}
        \addplot[color=rcasnap,mark=triangle*] coordinates {
          (0,6.445432) (1,4.896680) (2,4.464991) (3,4.121568)
          (4,4.010871) (5,4.040032) (6,4.082093)
        };
        \addlegendentry{SnapKV}

      \nextgroupplot[
        title={Thinking disabled},
        ymin=3.0,
        ymax=12.2,
      ]
        \addplot[color=rcanative,mark=*] coordinates {
          (0,7.033215) (1,8.303545) (2,8.322468) (3,8.683841)
          (4,9.619635) (5,10.761382) (6,11.515435)
        };
        \addplot[color=rcasw,mark=square*] coordinates {
          (0,4.132116) (1,4.557221) (2,4.476366) (3,4.480680)
          (4,4.722469) (5,4.472765) (6,4.206769)
        };
        \addplot[color=rcasnap,mark=triangle*] coordinates {
          (0,5.347785) (1,4.855199) (2,4.390666) (3,4.101749)
          (4,3.926326) (5,3.811806) (6,3.610172)
        };
    \end{groupplot}
  \end{tikzpicture}
  \caption{Interval \acceptancelength versus generated-token
  position for Qwen3.5-4B drafting Qwen3.5-122B responses, with thinking enabled and disabled. Reasoning magnifies the late staleness gap, but the SnapKV-SW crossover persists even without it.}
  \label{fig:rca-qwen-thinking-position}
\end{figure*}

\section{Additional Staleness-Aware Compression Results}
\label{app:section4-results}

\subsection{\riftstat Sliding-Window Ratio}
\label{app:sw-ratio-additional}
This experiment uses the base Qwen3.5-4B draft without fine-tuning. Figure~\ref{fig:sw-ratio-additional} supplements Figure~\ref{fig:effect-sw-ratio} with five additional output-cap and KV-budget combinations, evaluated on the same precomputed Qwen3.5-122B responses at $T=0$, with draft length 11, block size 128, and observation window 64. The best observed ratios range from $0.3$ to $0.7$ across these settings, with modest gains over the default $\rho=0.5$. In each panel, the best intermediate ratio outperforms both fixed-64 SnapKV and pure SWA. Output length is a cap, so naturally shorter responses retain their original lengths; comparisons across panels change the output cap and/or KV budget.

\definecolor{swratioLine}{RGB}{31,119,180}
\definecolor{swratioBest}{RGB}{241,181,34}
\definecolor{swratioBestEdge}{RGB}{163,105,0}
\definecolor{swratioEndpoint}{RGB}{95,95,95}
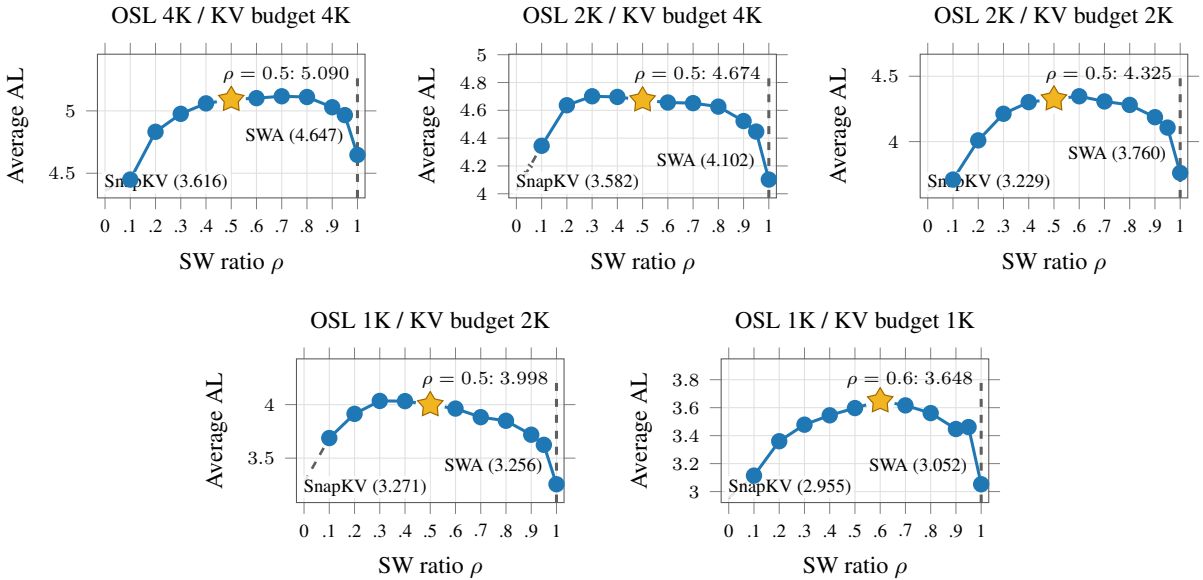
\begin{figure*}[!tbp]
\centering
\begin{minipage}[t]{0.32\linewidth}
\centering
\begin{tikzpicture}
\begin{axis}[
 width=\linewidth,height=0.68\linewidth,
 xmin=-0.03,xmax=1.03,ymin=4.302562562,ymax=5.454244842,
 xtick={0,0.1,0.2,0.3,0.4,0.5,0.6,0.7,0.8,0.9,1},
 xticklabels={0,.1,.2,.3,.4,.5,.6,.7,.8,.9,1},
 xlabel={SW ratio $\rho$},ylabel={Average AL},
 title={OSL 4K / KV budget 4K},title style={font=\small},
 grid=major,grid style={black!12},axis line style={black!65},tick align=outside,
 label style={font=\small},tick label style={font=\scriptsize},
 every axis plot/.append style={line width=1.15pt,mark size=2.5pt}]
\addplot[color=swratioEndpoint,dashed,forget plot] coordinates {(1,4.302562562) (1,5.304025414)};
\addplot[color=swratioLine,mark=none] coordinates {(0.10,4.449443780) (0.20,4.832247881) (0.30,4.976282204) (0.40,5.060033106) (0.50,5.090444832) (0.60,5.102074195) (0.70,5.117085682) (0.80,5.111168503) (0.90,5.029253842) (0.95,4.965956620) (1.00,4.646807179)};
\addplot[only marks,color=swratioLine,mark=*] coordinates {(0.10,4.449443780) (0.20,4.832247881) (0.30,4.976282204) (0.40,5.060033106) (0.60,5.102074195) (0.70,5.117085682) (0.80,5.111168503) (0.90,5.029253842) (0.95,4.965956620) (1.00,4.646807179)};
\node[circle,fill=white,minimum size=11pt,inner sep=0pt] at (axis cs:0.5,5.090444832) {};
\node[star,star points=5,star point ratio=1.8,fill=swratioBest,draw=swratioBestEdge,line width=0.45pt,minimum size=10pt,inner sep=0pt] at (axis cs:0.5,5.090444832) {};
\node[anchor=north east,font=\scriptsize] at (rel axis cs:0.98,0.98) {$\rho=0.5$: $5.090$};
\draw[color=swratioEndpoint,dashed,line width=1pt] (axis cs:0,4.362650333) -- (axis cs:0.1,4.449443780);
\draw[color=swratioEndpoint,line width=0.7pt] (axis cs:0.030000,4.410520257) -- (axis cs:0.048000,4.382479297);
\draw[color=swratioEndpoint,line width=0.7pt] (axis cs:0.046000,4.424407209) -- (axis cs:0.064000,4.396366249);
\node[anchor=south west,font=\scriptsize,fill=white,fill opacity=0.85,text opacity=1,inner sep=1pt] at (rel axis cs:0.015,0.025) {SnapKV (3.616)};
\node[anchor=south east,font=\scriptsize] at (axis cs:0.985,4.646807179) {SWA (4.647)};
\end{axis}
\end{tikzpicture}
\end{minipage}\hfill
\begin{minipage}[t]{0.32\linewidth}
\centering
\begin{tikzpicture}
\begin{axis}[
 width=\linewidth,height=0.68\linewidth,
 xmin=-0.03,xmax=1.03,ymin=3.970195876,ymax=5.003702188,
 xtick={0,0.1,0.2,0.3,0.4,0.5,0.6,0.7,0.8,0.9,1},
 xticklabels={0,.1,.2,.3,.4,.5,.6,.7,.8,.9,1},
 xlabel={SW ratio $\rho$},ylabel={Average AL},
 title={OSL 2K / KV budget 4K},title style={font=\small},
 grid=major,grid style={black!12},axis line style={black!65},tick align=outside,
 label style={font=\small},tick label style={font=\scriptsize},
 every axis plot/.append style={line width=1.15pt,mark size=2.5pt}]
\addplot[color=swratioEndpoint,dashed,forget plot] coordinates {(1,3.970195876) (1,4.868897017)};
\addplot[color=swratioLine,mark=none] coordinates {(0.10,4.344870182) (0.20,4.636627627) (0.30,4.701139471) (0.40,4.696270817) (0.50,4.673906191) (0.60,4.656263585) (0.70,4.651718079) (0.80,4.626670056) (0.90,4.522745016) (0.95,4.447209168) (1.00,4.102005377)};
\addplot[only marks,color=swratioLine,mark=*] coordinates {(0.10,4.344870182) (0.20,4.636627627) (0.30,4.701139471) (0.40,4.696270817) (0.60,4.656263585) (0.70,4.651718079) (0.80,4.626670056) (0.90,4.522745016) (0.95,4.447209168) (1.00,4.102005377)};
\node[circle,fill=white,minimum size=11pt,inner sep=0pt] at (axis cs:0.5,4.673906191) {};
\node[star,star points=5,star point ratio=1.8,fill=swratioBest,draw=swratioBestEdge,line width=0.45pt,minimum size=10pt,inner sep=0pt] at (axis cs:0.5,4.673906191) {};
\node[anchor=north east,font=\scriptsize] at (rel axis cs:0.98,0.98) {$\rho=0.5$: $4.674$};
\draw[color=swratioEndpoint,dashed,line width=1pt] (axis cs:0,4.024117945) -- (axis cs:0.1,4.344870182);
\draw[color=swratioEndpoint,line width=0.7pt] (axis cs:0.030000,4.161793133) -- (axis cs:0.048000,4.136629501);
\draw[color=swratioEndpoint,line width=0.7pt] (axis cs:0.046000,4.213113491) -- (axis cs:0.064000,4.187949859);
\node[anchor=south west,font=\scriptsize,fill=white,fill opacity=0.85,text opacity=1,inner sep=1pt] at (rel axis cs:0.015,0.025) {SnapKV (3.582)};
\node[anchor=south east,font=\scriptsize] at (axis cs:0.985,4.102005377) {SWA (4.102)};
\end{axis}
\end{tikzpicture}
\end{minipage}\hfill
\begin{minipage}[t]{0.32\linewidth}
\centering
\begin{tikzpicture}
\begin{axis}[
 width=\linewidth,height=0.68\linewidth,
 xmin=-0.03,xmax=1.03,ymin=3.569232641,ymax=4.668556087,
 xtick={0,0.1,0.2,0.3,0.4,0.5,0.6,0.7,0.8,0.9,1},
 xticklabels={0,.1,.2,.3,.4,.5,.6,.7,.8,.9,1},
 xlabel={SW ratio $\rho$},ylabel={Average AL},
 title={OSL 2K / KV budget 2K},title style={font=\small},
 grid=major,grid style={black!12},axis line style={black!65},tick align=outside,
 label style={font=\small},tick label style={font=\scriptsize},
 every axis plot/.append style={line width=1.15pt,mark size=2.5pt}]
\addplot[color=swratioEndpoint,dashed,forget plot] coordinates {(1,3.569232641) (1,4.525166072)};
\addplot[color=swratioLine,mark=none] coordinates {(0.10,3.709436211) (0.20,4.009371480) (0.30,4.212806250) (0.40,4.301826018) (0.50,4.325127339) (0.60,4.346725165) (0.70,4.307218838) (0.80,4.280624031) (0.90,4.186712962) (0.95,4.107009540) (1.00,3.760110897)};
\addplot[only marks,color=swratioLine,mark=*] coordinates {(0.10,3.709436211) (0.20,4.009371480) (0.30,4.212806250) (0.40,4.301826018) (0.60,4.346725165) (0.70,4.307218838) (0.80,4.280624031) (0.90,4.186712962) (0.95,4.107009540) (1.00,3.760110897)};
\node[circle,fill=white,minimum size=11pt,inner sep=0pt] at (axis cs:0.5,4.325127339) {};
\node[star,star points=5,star point ratio=1.8,fill=swratioBest,draw=swratioBestEdge,line width=0.45pt,minimum size=10pt,inner sep=0pt] at (axis cs:0.5,4.325127339) {};
\node[anchor=north east,font=\scriptsize] at (rel axis cs:0.98,0.98) {$\rho=0.5$: $4.325$};
\draw[color=swratioEndpoint,dashed,line width=1pt] (axis cs:0,3.626588647) -- (axis cs:0.1,3.709436211);
\draw[color=swratioEndpoint,line width=0.7pt] (axis cs:0.030000,3.672282265) -- (axis cs:0.048000,3.645516129);
\draw[color=swratioEndpoint,line width=0.7pt] (axis cs:0.046000,3.685537875) -- (axis cs:0.064000,3.658771739);
\node[anchor=south west,font=\scriptsize,fill=white,fill opacity=0.85,text opacity=1,inner sep=1pt] at (rel axis cs:0.015,0.025) {SnapKV (3.229)};
\node[anchor=south east,font=\scriptsize] at (axis cs:0.985,3.760110897) {SWA (3.760)};
\end{axis}
\end{tikzpicture}
\end{minipage}
\par\medskip
\begin{minipage}[t]{0.32\linewidth}
\centering
\begin{tikzpicture}
\begin{axis}[
 width=\linewidth,height=0.68\linewidth,
 xmin=-0.03,xmax=1.03,ymin=3.08466821,ymax=4.427647856,
 xtick={0,0.1,0.2,0.3,0.4,0.5,0.6,0.7,0.8,0.9,1},
 xticklabels={0,.1,.2,.3,.4,.5,.6,.7,.8,.9,1},
 xlabel={SW ratio $\rho$},ylabel={Average AL},
 title={OSL 1K / KV budget 2K},title style={font=\small},
 grid=major,grid style={black!12},axis line style={black!65},tick align=outside,
 label style={font=\small},tick label style={font=\scriptsize},
 every axis plot/.append style={line width=1.15pt,mark size=2.5pt}]
\addplot[color=swratioEndpoint,dashed,forget plot] coordinates {(1,3.084668210) (1,4.252476598)};
\addplot[color=swratioLine,mark=none] coordinates {(0.10,3.687599351) (0.20,3.913792493) (0.30,4.034485699) (0.40,4.032709048) (0.50,3.997923749) (0.60,3.963145005) (0.70,3.882729071) (0.80,3.849036865) (0.90,3.718803216) (0.95,3.624068675) (1.00,3.255946774)};
\addplot[only marks,color=swratioLine,mark=*] coordinates {(0.10,3.687599351) (0.20,3.913792493) (0.30,4.034485699) (0.40,4.032709048) (0.60,3.963145005) (0.70,3.882729071) (0.80,3.849036865) (0.90,3.718803216) (0.95,3.624068675) (1.00,3.255946774)};
\node[circle,fill=white,minimum size=11pt,inner sep=0pt] at (axis cs:0.5,3.997923749) {};
\node[star,star points=5,star point ratio=1.8,fill=swratioBest,draw=swratioBestEdge,line width=0.45pt,minimum size=10pt,inner sep=0pt] at (axis cs:0.5,3.997923749) {};
\node[anchor=north east,font=\scriptsize] at (rel axis cs:0.98,0.98) {$\rho=0.5$: $3.998$};
\draw[color=swratioEndpoint,dashed,line width=1pt] (axis cs:0,3.270991797) -- (axis cs:0.1,3.687599351);

\node[anchor=south west,font=\scriptsize,fill=white,fill opacity=0.85,text opacity=1,inner sep=1pt] at (rel axis cs:0.015,0.025) {SnapKV (3.271)};
\node[anchor=south east,font=\scriptsize] at (axis cs:0.985,3.255946774) {SWA (3.256)};
\end{axis}
\end{tikzpicture}
\end{minipage}\hspace{0.025\linewidth}
\begin{minipage}[t]{0.32\linewidth}
\centering
\begin{tikzpicture}
\begin{axis}[
 width=\linewidth,height=0.68\linewidth,
 xmin=-0.03,xmax=1.03,ymin=2.921436099,ymax=3.948917692,
 xtick={0,0.1,0.2,0.3,0.4,0.5,0.6,0.7,0.8,0.9,1},
 xticklabels={0,.1,.2,.3,.4,.5,.6,.7,.8,.9,1},
 xlabel={SW ratio $\rho$},ylabel={Average AL},
 title={OSL 1K / KV budget 1K},title style={font=\small},
 grid=major,grid style={black!12},axis line style={black!65},tick align=outside,
 label style={font=\small},tick label style={font=\scriptsize},
 every axis plot/.append style={line width=1.15pt,mark size=2.5pt}]
\addplot[color=swratioEndpoint,dashed,forget plot] coordinates {(1,2.921436099) (1,3.814898354)};
\addplot[color=swratioLine,mark=none] coordinates {(0.10,3.114115736) (0.20,3.360029082) (0.30,3.478312496) (0.40,3.545269827) (0.50,3.597021715) (0.60,3.648118733) (0.70,3.616232664) (0.80,3.561042534) (0.90,3.447344617) (0.95,3.461156424) (1.00,3.052477230)};
\addplot[only marks,color=swratioLine,mark=*] coordinates {(0.10,3.114115736) (0.20,3.360029082) (0.30,3.478312496) (0.40,3.545269827) (0.50,3.597021715) (0.70,3.616232664) (0.80,3.561042534) (0.90,3.447344617) (0.95,3.461156424) (1.00,3.052477230)};
\node[circle,fill=white,minimum size=11pt,inner sep=0pt] at (axis cs:0.6,3.648118733) {};
\node[star,star points=5,star point ratio=1.8,fill=swratioBest,draw=swratioBestEdge,line width=0.45pt,minimum size=10pt,inner sep=0pt] at (axis cs:0.6,3.648118733) {};
\node[anchor=north east,font=\scriptsize] at (rel axis cs:0.98,0.98) {$\rho=0.6$: $3.648$};
\draw[color=swratioEndpoint,dashed,line width=1pt] (axis cs:0,2.954929176) -- (axis cs:0.1,3.114115736);

\node[anchor=south west,font=\scriptsize,fill=white,fill opacity=0.85,text opacity=1,inner sep=1pt] at (rel axis cs:0.015,0.025) {SnapKV (2.955)};
\node[anchor=south east,font=\scriptsize] at (axis cs:0.985,3.052477230) {SWA (3.052)};
\end{axis}
\end{tikzpicture}
\end{minipage}
\caption{Additional sliding-window ratio ablations with the same base draft model, requests, and evaluation settings as Figure~\ref{fig:effect-sw-ratio}. Panel titles specify the output-length cap and KV budget. Stars mark $\rho=0.5$ in the first four panels and the optimum $\rho=0.6$ in the 1K-output, 1K-budget panel (AL $3.648$, versus $3.597$ at $\rho=0.5$); the zero-labeled endpoint denotes fixed-64 SnapKV, and $\rho=1$ denotes pure SWA. Broken diagonals indicate omitted AL range.}
\label{fig:sw-ratio-additional}
\label{fig:sw-ratio-1k}
\end{figure*}

\subsection{Rescoring Overhead Estimation}\label{app:rescoring-overhead}
Figure~\ref{fig:effect-dynamic-refresh} in the main paper projects the added rescoring overhead for Qwen3.5-4B with 100K ISL and 4K OSL. For a given refresh interval, we begin by calculating the drafter's AL under \riftdyn with the corresponding refresh interval. Then, we run the speculative decoding loop in vLLM with the corresponding forced AL and a fixed 8K draft KV cache to measure decode latency. We add the estimated vLLM rescoring time for the given interval, and report the added time as a percentage of measured decode latency.

\subsection{Block-Size Ablation}
\label{app:block-size-ablation}
As described in Section~\ref{sec:staleness-aware-kv-compression}, to accommodate real-world serving scenarios, we work with block-level KV cache compression.
We examine the effect of block size on the draft's performance. We consider Qwen3.5-4B with draft length 11, drafting for Qwen3.5-122B using the SPEED-Bench dataset, 100K and 32K throughput splits. In each setting, we apply \riftstat with the corresponding block size.
We consider two generation temperatures and sweep various block sizes. As seen in Figure~\ref{fig:block-size-ablation}, when the block size is significantly smaller than the overall KV cache, the effect is negligible. However, each setting has a "critical block size" beyond which the decision space is too coarse and the draft's performance degrades.

\subsection{Budget, Context, and Cross-Family Comparisons}

We supplement the results in Section~\ref{sec:results} with additional cache-budget sweeps in various settings.

\textbf{Additional Qwen Family Results}
Table~\ref{tab:section4-qwen-workload-budget-appendix} expands the Qwen3.5
results across context length, output entropy, temperature, and cache budget.
The gains from allocating cache capacity to recent tokens and refreshing the
remote selection remain consistent throughout these settings.
Table~\ref{tab:trainable-snapkv-qwen08b-base-relative} presents additional results with a Qwen3.5-0.8B draft model. As can be seen, the performance extends to smaller models.

\paragraph{Additional model families and cross-family drafting}
\label{app:cross-family-drafting}
We extend our experiments beyond the Qwen3.5 family and the pairs considered in Section~\ref{sec:results}.
We provide several settings:
\begin{enumerate}
    \item Nemotron3 family - Due to their sizes, we consider Nemotron3-Super and Nemotron3-Ultra as target models, with Nemotron3-Nano and Nemotron3.5-Lightning serving as their draft models. Additionally, we consider a self-speculation setting using Nemotron3.5-Lightning. Numerics are given in Table~\ref{tab:section4-nemotron-budget-appendix}.
    \item Additional Model families - Llama3.1 (8B drafting for 70B) serving as a full-attention example, and GPTOSS (20B drafting for 120B) serving as a mixed-attention example. Numerics are given in Table~\ref{tab:section4-crossmodel-budget-appendix}.
    \item We provide additional cross-family drafting results. First, we consider additional draft models for the DeepSeek-V4 targets (Flash and Pro) in Tables~\ref{tab:deepseek-v4-pro} and \ref{tab:deepseek-v4-flash} We consider Qwen3.5-4B, Qwen3.5-0.8B and Llama3.2-1B. Second, we consider a Kimi-K3 target model and a Qwen3.5-4B draft in Table~\ref{tab:trainable-snapkv-cross-model}.

\end{enumerate}
It can be seen that the performance of \rift is maintained across all considered settings. It can also be seen that some setups benefit less from \riftstat, showing that similar performance can be obtained with SWA at the same budget (see, for example, Nemotron3-Nano drafting for Nemotron3-Ultra or Super in Table~\ref{tab:section4-nemotron-budget-appendix}). However, the benefit of dynamically rescoring and updating the KV cache during decode dramatically increases the draft's performance, regardless of the specific setup considered.

\subsection{Additional Prefill Selectors}
Table~\ref{tab:static-selectors-sw-ratio-full} reports the absolute ALs underlying the relative gains in Table~\ref{tab:static-selectors-sw-ratio}.

{\color{introrevision}
\par\medskip\refstepcounter{subsection}
\label{app:effect-draft-length}
\noindent\textbf{\thesubsection. Effect of draft length}\par\smallskip
We evaluate the effect of draft length on performance in Full-KV, SnapKV, SW, \riftstat, and \riftdyn at $T\in\{0,1\}$.
As seen in Figure~\ref{fig:effect-draft-length}, the performance gap persists across draft lengths: \riftstat and \riftdyn consistently outperform SnapKV, with \riftdyn closest to Full-KV.
The relative AL gain of \riftdyn over SnapKV grows from $17.7\%$ to $60.0\%$ at $T=0$ and from $17.9\%$ to $54.8\%$ at $T=1$ as draft length increases.
}
\definecolor{dlfull}{RGB}{28,28,28}
\definecolor{dlsw}{RGB}{135,135,135}
\definecolor{dlsnap}{RGB}{205,164,0}
\definecolor{dlstat}{RGB}{62,132,0}
\definecolor{dldyn}{RGB}{0,145,105}
\begin{figure*}[!tbp]
\centering
\begin{tikzpicture}
\begin{groupplot}[
 group style={group size=2 by 1,horizontal sep=0.9cm},
 width=0.37\textwidth,height=0.26\textwidth,
 xmin=2.5,xmax=17.5,ymin=2.4,ymax=7.8,
 xtick={3,5,7,11,13,15,17},ytick={3,4,5,6,7},
 xlabel={Draft length},ylabel={Mean AL},
 tick label style={font=\scriptsize},label style={font=\small},
 title style={font=\small},grid=major,major grid style={black!13},
 axis line style={black!65},tick align=outside,
 legend style={draw=none,font=\scriptsize,/tikz/every even column/.append style={column sep=7pt}},
 legend columns=5,
]
\nextgroupplot[title={$T=0$},legend to name=dlsharedlegend]
\addplot[dlfull,dashed,mark=*,line width=1pt,mark size=1.7pt,mark options={solid}] coordinates {
(3,3.320988136) (5,4.424263649) (7,5.281202653) (11,6.465772877) (13,6.885293299) (15,7.221202461) (17,7.491700956)
};
\addlegendentry{Full-KV}
\addplot[dlsw,dashdotted,mark=square*,line width=1pt,mark size=1.7pt,mark options={solid}] coordinates {
(3,3.026262046) (5,3.860890629) (7,4.459546544) (11,5.230172198) (13,5.487535953) (15,5.687957856) (17,5.851261575)
};
\addlegendentry{SW}
\addplot[dlsnap,dotted,mark=triangle*,line width=1pt,mark size=1.7pt,mark options={solid}] coordinates {
(3,2.753867845) (5,3.339345233) (7,3.704811761) (11,4.104163034) (13,4.218715775) (15,4.303605359) (17,4.364034612)
};
\addlegendentry{SnapKV}
\addplot[dlstat,solid,mark=diamond*,line width=1.5pt,mark size=1.7pt,mark options={solid}] coordinates {
(3,3.138394206) (5,4.063354382) (7,4.741880667) (11,5.627984653) (13,5.925352647) (15,6.162706031) (17,6.347137322)
};
\addlegendentry{\riftstat}
\addplot[dldyn,solid,mark=star,line width=1.6pt,mark size=1.7pt,mark options={solid}] coordinates {
(3,3.240998933) (5,4.264021220) (7,5.038243920) (11,6.091776809) (13,6.455345480) (15,6.750361244) (17,6.983607796)
};
\addlegendentry{\riftdyn}
\nextgroupplot[title={$T=1$},ylabel={}]
\addplot[dlfull,dashed,mark=*,line width=1pt,mark size=1.7pt,mark options={solid}] coordinates {
(3,3.227707272) (5,4.235715058) (7,4.986661579) (11,5.977760527) (13,6.308218080) (15,6.566056476) (17,6.771993346)
};
\addplot[dlsw,dashdotted,mark=square*,line width=1pt,mark size=1.7pt,mark options={solid}] coordinates {
(3,2.892961167) (5,3.614084946) (7,4.106417017) (11,4.698337195) (13,4.880884775) (15,5.020581044) (17,5.127907244)
};
\addplot[dlsnap,dotted,mark=triangle*,line width=1pt,mark size=1.7pt,mark options={solid}] coordinates {
(3,2.631322422) (5,3.137840932) (7,3.439902313) (11,3.752867097) (13,3.837462939) (15,3.896805789) (17,3.940809176)
};
\addplot[dlstat,solid,mark=diamond*,line width=1.5pt,mark size=1.7pt,mark options={solid}] coordinates {
(3,2.985768594) (5,3.774866046) (7,4.322733660) (11,4.990343026) (13,5.198120273) (15,5.357634244) (17,5.480011690)
};
\addplot[dldyn,solid,mark=star,line width=1.6pt,mark size=1.7pt,mark options={solid}] coordinates {
(3,3.102104682) (5,3.994937726) (7,4.638034598) (11,5.461719623) (13,5.729418744) (15,5.937938808) (17,6.101899841)
};
\end{groupplot}
\node[anchor=north,yshift=-1.05cm] at ($(group c1r1.south)!0.5!(group c2r1.south)$) {\pgfplotslegendfromname{dlsharedlegend}};
\end{tikzpicture}
\caption{\textcolor{introrevision}{Effect of draft length for Qwen3.5-4B drafting for Qwen3.5-122B-A10B on low-entropy SPEED-Bench 32K prompts, with a 4K output cap. Compressed methods use an 8K KV budget and block size 128; Full-KV retains the full context. Each point is mean per-request AL.}}
\label{fig:effect-draft-length}
\end{figure*}
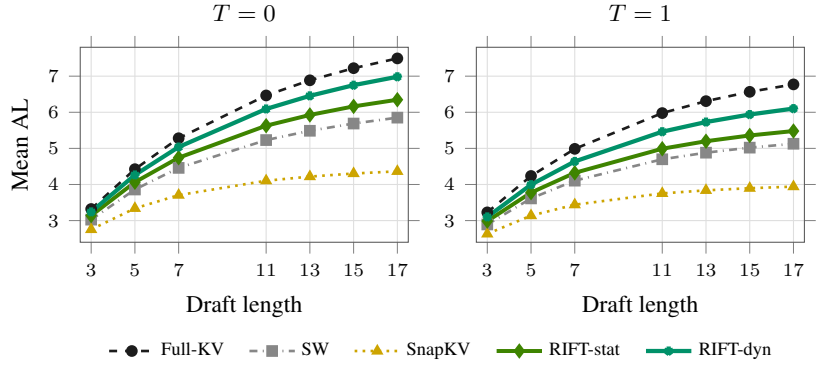

{
\par\medskip\refstepcounter{subsection}
\label{app:al-vs-osl}
\noindent\textbf{\thesubsection. AL vs. OSL}\par\smallskip
We evaluate how acceptance changes with OSL on
SPEED-Bench 32K low-entropy data. We force each target response to meet OSL
shown in Figure~\ref{fig:al-vs-osl}. In this forced-output setting, the advantage of \riftstat
over SWA diminishes as OSL grows: their mean-AL gap narrows from $0.42$ to
$0.09$ at $T=0$ and from $0.29$ to $0.03$ at $T=1$ between 2K and 64K OSL.
In contrast, the advantage of the decode-level methods grows with OSL. 
We note that forcing high OSLs produces highly repetitive generated text, which may affect both the measured ALs and the relative gaps between methods. Thus, the observed trends should therefore be interpreted as results for forced continuations rather than naturally long, meaningful responses.
}
\definecolor{oslsw}{RGB}{135,135,135}
\definecolor{oslsnap}{RGB}{205,164,0}
\definecolor{oslstat}{RGB}{62,132,0}
\definecolor{osldyn}{RGB}{0,145,105}
\definecolor{oslada}{RGB}{145,70,180}
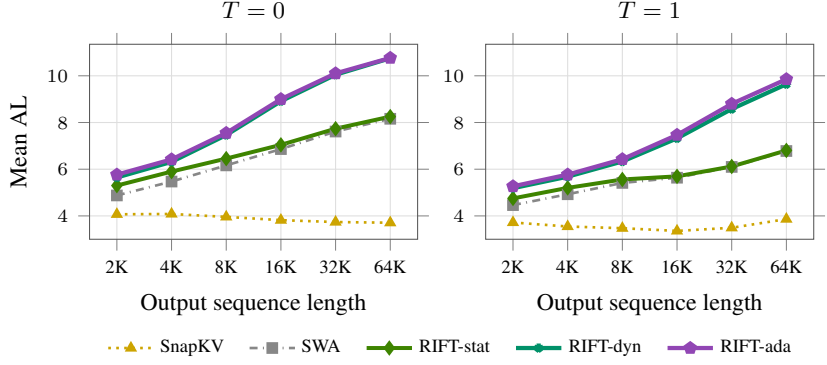
\begin{figure*}[!tbp]
  \centering
  \begin{tikzpicture}
    \begin{groupplot}[
      group style={group size=2 by 1,horizontal sep=0.9cm},
      width=0.37\textwidth,
      height=0.26\textwidth,
      xmode=log,
      log basis x=2,
      xtick={2048,4096,8192,16384,32768,65536},
      xticklabels={2K,4K,8K,16K,32K,64K},
      xlabel={Output sequence length},
      ymin=3.0,
      ymax=11.35,
      grid=major,
      major grid style={black!13},
      axis line style={black!65},tick align=outside,
      tick label style={font=\scriptsize},
      label style={font=\small},
      title style={font=\small},
      legend style={font=\scriptsize,draw=none,/tikz/every even column/.append style={column sep=7pt}},
    ]
    \nextgroupplot[title={${T=0}$},ylabel={Mean AL},legend columns=5,legend to name=alvsosllegend]
      \addplot[oslsnap,dotted,mark=triangle*,line width=1pt,mark size=1.7pt,mark options={solid}] coordinates {(2048,4.069468) (4096,4.083320) (8192,3.958631) (16384,3.818434) (32768,3.737759) (65536,3.708486)};
      \addlegendentry{SnapKV}
      \addplot[oslsw,dashdotted,mark=square*,line width=1pt,mark size=1.7pt,mark options={solid}] coordinates {(2048,4.876576) (4096,5.472207) (8192,6.163831) (16384,6.862641) (32768,7.613709) (65536,8.159586)};
      \addlegendentry{SWA}
      \addplot[oslstat,solid,mark=diamond*,line width=1.5pt,mark size=1.7pt,mark options={solid}] coordinates {(2048,5.299513) (4096,5.898385) (8192,6.455154) (16384,7.043546) (32768,7.735406) (65536,8.252521)};
      \addlegendentry{RIFT-stat}
      \addplot[osldyn,solid,mark=star,line width=1.6pt,mark size=1.7pt,mark options={solid}] coordinates {(2048,5.645723) (4096,6.313965) (8192,7.462787) (16384,8.911815) (32768,10.046550) (65536,10.752863)};
      \addlegendentry{RIFT-dyn}
      \addplot[oslada,solid,mark=pentagon*,line width=1.6pt,mark size=1.7pt,mark options={solid}] coordinates {(2048,5.766641) (4096,6.419679) (8192,7.549979) (16384,8.999589) (32768,10.102989) (65536,10.765052)};
      \addlegendentry{RIFT-ada}
    \nextgroupplot[title={${T=1}$},ylabel={}]
      \addplot[oslsnap,dotted,mark=triangle*,line width=1pt,mark size=1.7pt,mark options={solid}] coordinates {(2048,3.718822) (4096,3.546426) (8192,3.473530) (16384,3.357432) (32768,3.491924) (65536,3.863363)};
      \addplot[oslsw,dashdotted,mark=square*,line width=1pt,mark size=1.7pt,mark options={solid}] coordinates {(2048,4.463161) (4096,4.927721) (8192,5.413297) (16384,5.636079) (32768,6.098114) (65536,6.779609)};
      \addplot[oslstat,solid,mark=diamond*,line width=1.5pt,mark size=1.7pt,mark options={solid}] coordinates {(2048,4.749839) (4096,5.200476) (8192,5.561155) (16384,5.694630) (32768,6.116437) (65536,6.804706)};
      \addplot[osldyn,solid,mark=star,line width=1.6pt,mark size=1.7pt,mark options={solid}] coordinates {(2048,5.187509) (4096,5.680849) (8192,6.331589) (16384,7.321881) (32768,8.576438) (65536,9.640246)};
      \addplot[oslada,solid,mark=pentagon*,line width=1.6pt,mark size=1.7pt,mark options={solid}] coordinates {(2048,5.266753) (4096,5.768604) (8192,6.427755) (16384,7.468319) (32768,8.798236) (65536,9.853755)};
    \end{groupplot}
    \node[anchor=north,yshift=-1.05cm] at ($(group c1r1.south)!0.5!(group c2r1.south)$) {\pgfplotslegendfromname{alvsosllegend}};
  \end{tikzpicture}
  \caption{Mean per-request acceptance length
  versus forced output sequence length for Qwen3.5-4B drafting for
  Qwen3.5-122B-A10B on SPEED-Bench 32K low-entropy prompts. All compressed
  methods use an 8K KV budget and draft length 11.}
  \label{fig:al-vs-osl}
\end{figure*}

\begin{table*}[!tbp]
  \centering
\footnotesize
  \setlength{\tabcolsep}{4.5pt}
  \begin{tabular}{llllccccc}
    \toprule
    ISL & Entropy & $T$ & Budget & SW & SnapKV & \riftstat & \riftdyn & \riftada \\
    \midrule
    \multirow{9}{*}{32K}
      & \multirow{3}{*}{Low} & \multirow{3}{*}{0}
      & 2K & 64.8\% & 49.7\% & 70.3\% & 79.5\% & 81.5\% \\
      & & & 4K & 71.3\% & 54.5\% & 77.7\% & 88.5\% & 90.2\% \\
      & & & 8K & 80.9\% & 63.0\% & 86.9\% & 94.6\% & 95.8\% \\
      \cmidrule(lr){2-9}
      & \multirow{3}{*}{Low} & \multirow{3}{*}{1}
      & 2K & 61.8\% & 47.1\% & 65.9\% & 74.7\% & 77.0\% \\
      & & & 4K & 68.0\% & 52.0\% & 72.5\% & 84.1\% & 86.3\% \\
      & & & 8K & 78.6\% & 61.8\% & 83.4\% & 91.7\% & 92.8\% \\
      \cmidrule(lr){2-9}
      & \multirow{3}{*}{High} & \multirow{3}{*}{0}
      & 2K & 79.1\% & 61.9\% & 81.8\% & 88.5\% & 90.1\% \\
      & & & 4K & 84.0\% & 64.4\% & 88.8\% & 94.1\% & 94.7\% \\
      & & & 8K & 85.9\% & 66.9\% & 93.1\% & 97.6\% & 97.7\% \\
    \midrule
    \multirow{6}{*}{100K}
      & \multirow{3}{*}{Low} & \multirow{3}{*}{0}
      & 2K & 64.1\% & 50.7\% & 72.1\% & 81.7\% & 84.7\% \\
      & & & 4K & 75.5\% & 55.5\% & 77.8\% & 90.3\% & 92.4\% \\
      & & & 8K & 81.8\% & 59.7\% & 85.6\% & 95.5\% & 96.0\% \\
      \cmidrule(lr){2-9}
      & \multirow{3}{*}{High} & \multirow{3}{*}{0}
      & 2K & 72.7\% & 60.0\% & 80.2\% & 86.9\% & 88.6\% \\
      & & & 4K & 81.1\% & 62.5\% & 86.7\% & 92.1\% & 93.2\% \\
      & & & 8K & 84.7\% & 65.2\% & 90.7\% & 95.9\% & 96.3\% \\
    \bottomrule
  \end{tabular}
\caption{retained AL (\%) of full-KV, Qwen3.5-4B draft with, Qwen3.5-122B target across SPEED-Bench workloads (ISL and entropy category) and cache budgets. The absolute full-KV ALs for 32K low $T=0/1$, 32K high $T=0$, 100K low $T=0$,
and 100K high $T=0$ are $6.466/5.978$, $4.617$, $7.472$, and $4.506$, respectively. All settings use draft length 11 and maximum output length 4K. 
}
\label{tab:section4-qwen-workload-budget-appendix}
\end{table*}

\begin{table*}[!tbp]
  \centering
\footnotesize
  \setlength{\tabcolsep}{4.5pt}
  \begin{tabular}{llccrrrrr}
    \toprule
    ISL & Entropy & $T$
      & SW & SnapKV & \riftstat & \riftdyn & \riftada \\
    \midrule
    \multirow{4}{*}{32K}
      & Low  & 0   & $85.5\%$ & $67.3\%$ & $89.0\%$ & $95.9\%$ & $98.1\%$ \\
      & Low  & 1   & $85.5\%$ & $68.2\%$ & $86.9\%$ & $94.2\%$ & $96.4\%$ \\
      & High & 0   & $91.5\%$ & $75.0\%$ & $95.6\%$ & $97.9\%$ & $98.6\%$ \\
      & High & 1   & $92.6\%$ & $77.7\%$ & $94.9\%$ & $97.1\%$ & $97.9\%$ \\
    \midrule
    \multirow{3}{*}{100K}
      & Low  & 0   & $82.6\%$ & $61.0\%$ & $87.5\%$ & $94.6\%$ & $97.4\%$ \\
      & Low  & 1   & $84.6\%$ & $62.0\%$ & $86.9\%$ & $92.7\%$ & $97.1\%$ \\
      & High & 0   & $90.6\%$ & $71.9\%$ & $94.6\%$ & $96.3\%$ & $97.7\%$ \\
    \bottomrule
  \end{tabular}
  \caption{Base Qwen3.5-0.8B draft performance at an 8K cache budget,
  expressed as a percentage of the matching Full KV AL for
  Qwen3.5-122B target responses. The absolute full-KV ALs for 32K low
  $T=0/1$ and high $T=0/1$ are $4.807/4.148$ and $3.272/2.836$; for 100K low
  $T=0/1$ and high $T=0$, they are $5.763/4.748$ and $3.225$.}
  \label{tab:trainable-snapkv-qwen08b-base-relative}
\end{table*}

\begin{table*}[!t]
  \centering
\footnotesize
  \setlength{\tabcolsep}{4pt}
  \begin{tabular}{lllccccc}
    \toprule
    Draft & Target & $T$ & Budget & SW & SnapKV & \riftstat & \riftdyn \\
    \midrule
    \multirow{3}{*}{Nano} & \multirow{3}{*}{Super} & \multirow{3}{*}{0}
      & 2K & 53.0\% & 41.8\% & 52.6\% & 62.1\% \\
      & & & 4K & 57.1\% & 45.1\% & 57.2\% & 73.2\% \\
      & & & 8K & 63.4\% & 52.8\% & 63.9\% & 83.4\% \\
    \midrule
    \multirow{3}{*}{Nano} & \multirow{3}{*}{Ultra} & \multirow{3}{*}{0}
      & 2K & 69.1\% & 56.0\% & 69.9\% & 76.6\% \\
      & & & 4K & 74.4\% & 60.5\% & 75.2\% & 85.9\% \\
      & & & 8K & 81.3\% & 68.4\% & 82.8\% & 93.2\% \\
    \midrule
    \multirow{6}{*}{Lightning} & \multirow{6}{*}{Super}
      & \multirow{3}{*}{0} & 2K & 52.6\% & 42.1\% & 53.3\% & 66.9\% \\
      & & & 4K & 56.3\% & 45.1\% & 56.9\% & 78.2\% \\
      & & & 8K & 62.2\% & 51.9\% & 62.2\% & 90.7\% \\
      \cmidrule(lr){3-8}
      & & \multirow{3}{*}{1} & 2K & 56.1\% & 44.8\% & 57.1\% & 68.4\% \\
      & & & 4K & 60.7\% & 48.6\% & 61.7\% & 78.7\% \\
      & & & 8K & 68.2\% & 56.2\% & 68.1\% & 90.2\% \\
    \midrule
    \multirow{6}{*}{Lightning} & \multirow{6}{*}{Ultra}
      & \multirow{3}{*}{0} & 2K & 67.0\% & 54.6\% & 69.6\% & 77.2\% \\
      & & & 4K & 72.7\% & 59.3\% & 74.0\% & 84.8\% \\
      & & & 8K & 79.9\% & 67.4\% & 80.7\% & 92.6\% \\
      \cmidrule(lr){3-8}
      & & \multirow{3}{*}{1} & 2K & 66.1\% & 53.8\% & 69.8\% & 76.7\% \\
      & & & 4K & 71.8\% & 58.1\% & 74.1\% & 83.9\% \\
      & & & 8K & 79.5\% & 66.2\% & 80.8\% & 91.6\% \\
    \midrule
    \multirow{6}{*}{Lightning} & \multirow{6}{*}{Lightning}
      & \multirow{3}{*}{0} & 2K & 43.0\% & 35.8\% & 46.8\% & 57.6\% \\
      & & & 4K & 49.7\% & 40.5\% & 54.1\% & 70.2\% \\
      & & & 8K & 57.3\% & 48.3\% & 67.2\% & 82.6\% \\
      \cmidrule(lr){3-8}
      & & \multirow{3}{*}{1} & 2K & 35.3\% & 29.2\% & 40.2\% & 51.2\% \\
      & & & 4K & 41.4\% & 33.4\% & 47.1\% & 64.2\% \\
      & & & 8K & 49.5\% & 40.2\% & 60.9\% & 77.5\% \\
    \bottomrule
  \end{tabular}
  \caption{Full-KV AL retention in Nemotron3-family, low-entropy SPEED-Bench
  32K split. The absolute full-KV ALs are $6.944$ (Nano $\rightarrow$ Super,
  $T=0$), $4.994$ (Nano $\rightarrow$ Ultra, $T=0$), $6.647/5.520$
  (Lightning $\rightarrow$ Super, $T=0/1$), $5.394/4.859$ (Lightning
  $\rightarrow$ Ultra, $T=0/1$), and $10.009/11.222$ (Lightning
  self-speculation, $T=0/1$).}
  \label{tab:section4-nemotron-budget-appendix}
\end{table*}

\begin{table*}[!t]
  \centering
\footnotesize
  \setlength{\tabcolsep}{4.5pt}
  \begin{tabular}{llcccc}
    \toprule
    Draft $\rightarrow$ target & Budget & SW & SnapKV & \riftstat & \riftdyn \\
    \midrule
    \multirow{3}{*}{Llama-3.1-8B $\rightarrow$ Llama-3.1-70B}
      & 2K & 26.6\% & 53.7\% & 64.3\% & 70.5\% \\
      & 4K & 27.6\% & 58.1\% & 70.1\% & 79.7\% \\
      & 8K & 28.6\% & 66.1\% & 77.4\% & 88.0\% \\
    \midrule
    \multirow{3}{*}{GPT-OSS-20B $\rightarrow$ GPT-OSS-120B}
      & 2K & 62.3\% & 32.0\% & 56.4\% & 67.7\% \\
      & 4K & 65.5\% & 32.6\% & 64.9\% & 76.8\% \\
      & 8K & 69.6\% & 34.0\% & 71.4\% & 87.8\% \\
    \bottomrule
  \end{tabular}
  \caption{Llama3.1 and GPT-OSS cache-budget sweeps on the $T=0$,
  low-entropy SPEED-Bench 32K workload. Values are \acceptancelength as a
  percentage of the matching full-KV AL. The absolute full-KV AL is $7.849$
  for Llama and $5.746$ for GPT-OSS. All settings use draft length 11 and
  maximum output length 4K.}
  \label{tab:section4-crossmodel-budget-appendix}
\end{table*}

\begin{table*}[!tbp]
  \centering
\footnotesize
  \setlength{\tabcolsep}{3pt}
  \begin{subtable}[t]{0.49\linewidth}
    \centering
    \caption{32K ISL}
    \label{tab:deepseek-v4-pro-32k}

    \begin{tabular}{llccc}
      \toprule
      Draft & Budget & SnapKV & \riftstat & \riftdyn \\
      \midrule
      \multirow{4}{*}{Qwen3.5-0.8B}
        & 2K  & 48.4\% & 63.3\% & 78.8\% \\
        & 4K  & 54.7\% & 69.9\% & 87.6\% \\
        & 8K  & 63.2\% & 77.6\% & 93.4\% \\
        & 16K & 70.5\% & 84.7\% & 97.5\% \\
      \cmidrule(lr){2-5}
      \multirow{4}{*}{Qwen3.5-4B}
        & 2K  & 49.3\% & 63.5\% & 78.9\% \\
        & 4K  & 54.7\% & 70.0\% & 88.2\% \\
        & 8K  & 63.7\% & 79.0\% & 94.1\% \\
        & 16K & 70.1\% & 85.2\% & 98.0\% \\
      \cmidrule(lr){2-5}
      \multirow{4}{*}{Llama3.2-1B}
        & 2K  & 51.6\% & 66.8\% & 70.6\% \\
        & 4K  & 54.1\% & 71.3\% & 75.9\% \\
        & 8K  & 59.8\% & 77.3\% & 82.1\% \\
        & 16K & 67.4\% & 84.7\% & 90.5\% \\
      \bottomrule
    \end{tabular}
  \end{subtable}\hfill%
  \begin{subtable}[t]{0.49\linewidth}
    \centering
    \caption{100K ISL}
    \label{tab:deepseek-v4-pro-100k}

    \begin{tabular}{llccc}
      \toprule
      Draft & Budget & SnapKV & \riftstat & \riftdyn \\
      \midrule
      \multirow{4}{*}{Qwen3.5-0.8B}
        & 2K  & 57.9\% & 76.1\% & 84.6\% \\
        & 4K  & 63.1\% & 81.2\% & 92.4\% \\
        & 8K  & 66.8\% & 88.9\% & 96.0\% \\
        & 16K & 69.4\% & 92.2\% & 97.7\% \\
      \cmidrule(lr){2-5}
      \multirow{4}{*}{Qwen3.5-4B}
        & 2K  & 56.5\% & 74.2\% & 83.3\% \\
        & 4K  & 62.3\% & 79.5\% & 91.5\% \\
        & 8K  & 65.9\% & 87.6\% & 95.2\% \\
        & 16K & 68.5\% & 91.5\% & 97.1\% \\
      \cmidrule(lr){2-5}
      \multirow{4}{*}{Llama3.2-1B}
        & 2K  & 56.1\% & 77.5\% & 80.4\% \\
        & 4K  & 58.6\% & 82.2\% & 85.0\% \\
        & 8K  & 61.8\% & 89.4\% & 91.4\% \\
        & 16K & 65.0\% & 92.3\% & 94.4\% \\
      \bottomrule
    \end{tabular}
  \end{subtable}
  \caption{DeepSeek V4 Pro with trained external drafters on the
  $T=0$, low-entropy SPEED-Bench workloads. For Qwen3.5-0.8B, Qwen3.5-4B, and Llama3.2-1B,
  respectively, the absolute full-KV ALs are $4.343/5.389/3.969$ at 32K and
  $4.263/5.427/3.674$ at 100K.}
  \label{tab:deepseek-v4-pro}
\end{table*}

\begin{table*}[!tbp]
  \centering
\footnotesize
  \setlength{\tabcolsep}{3pt}
  \begin{subtable}[t]{0.49\linewidth}
    \centering
    \caption{32K ISL}
    \label{tab:deepseek-v4-flash-32k}

    \begin{tabular}{llccc}
      \toprule
      Draft & Budget & SnapKV & \riftstat & \riftdyn \\
      \midrule
      \multirow{4}{*}{Qwen3.5-0.8B}
        & 2K  & 40.9\% & 54.4\% & 75.0\% \\
        & 4K  & 50.6\% & 62.6\% & 85.5\% \\
        & 8K  & 59.8\% & 72.4\% & 92.7\% \\
        & 16K & 67.9\% & 79.3\% & 97.2\% \\
      \cmidrule(lr){2-5}
      \multirow{4}{*}{Qwen3.5-4B}
        & 2K  & 42.7\% & 55.3\% & 75.1\% \\
        & 4K  & 51.3\% & 63.1\% & 86.4\% \\
        & 8K  & 61.3\% & 74.5\% & 93.7\% \\
        & 16K & 68.9\% & 80.7\% & 97.8\% \\
      \cmidrule(lr){2-5}
      \multirow{4}{*}{Llama3.2-1B}
        & 2K  & 46.6\% & 59.3\% & 64.1\% \\
        & 4K  & 50.1\% & 64.9\% & 71.2\% \\
        & 8K  & 56.7\% & 73.6\% & 79.6\% \\
        & 16K & 65.7\% & 81.0\% & 89.4\% \\
      \bottomrule
    \end{tabular}
  \end{subtable}\hfill%
  \begin{subtable}[t]{0.49\linewidth}
    \centering
    \caption{100K ISL}
    \label{tab:deepseek-v4-flash-100k}

    \begin{tabular}{llccc}
      \toprule
      Draft & Budget & SnapKV & \riftstat & \riftdyn \\
      \midrule
      \multirow{4}{*}{Qwen3.5-0.8B}
        & 2K  & 55.0\% & 62.2\% & 78.7\% \\
        & 4K  & 63.4\% & 71.1\% & 89.0\% \\
        & 8K  & 70.4\% & 86.1\% & 95.0\% \\
        & 16K & 76.3\% & 92.7\% & 97.3\% \\
      \cmidrule(lr){2-5}
      \multirow{4}{*}{Qwen3.5-4B}
        & 2K  & 56.0\% & 62.5\% & 79.8\% \\
        & 4K  & 65.0\% & 71.2\% & 89.6\% \\
        & 8K  & 71.8\% & 86.4\% & 95.1\% \\
        & 16K & 76.4\% & 92.9\% & 97.6\% \\
      \cmidrule(lr){2-5}
      \multirow{4}{*}{Llama3.2-1B}
        & 2K  & 49.2\% & 61.4\% & 64.8\% \\
        & 4K  & 54.0\% & 69.6\% & 75.1\% \\
        & 8K  & 60.2\% & 84.4\% & 87.9\% \\
        & 16K & 67.0\% & 91.5\% & 93.6\% \\
      \bottomrule
    \end{tabular}
  \end{subtable}
  \caption{DeepSeek V4 Flash with trained external drafters on the
  $T=0$, low-entropy SPEED-Bench workloads. For Qwen3.5-0.8B, Qwen3.5-4B, and Llama3.2-1B,
  respectively, the absolute full-KV ALs are $4.830/5.714/4.401$ at 32K and
  $5.308/6.338/4.896$ at 100K.}
  \label{tab:deepseek-v4-flash}
\end{table*}

\begin{table*}[!tbp]
  \centering
\footnotesize
  \setlength{\tabcolsep}{4.2pt}

  \begin{tabular}{lccccccccc}
    \toprule
    & \multicolumn{3}{c}{\riftstat} & \multicolumn{3}{c}{\riftdyn} & \multicolumn{3}{c}{\riftada} \\
    \cmidrule(lr){2-4}\cmidrule(lr){5-7}\cmidrule(lr){8-10}
    $T$ & 2K & 4K & 8K & 2K & 4K & 8K & 2K & 4K & 8K \\
    \midrule
    0 & 60.5\% & 66.0\% & 72.7\% & 84.3\% & 93.4\% & 97.1\% & 88.4\% & 94.8\% & 98.3\% \\
    1 & 56.6\% & 62.5\% & 69.3\% & 81.5\% & 91.6\% & 96.1\% & 87.2\% & 94.7\% & 98.2\% \\
    \bottomrule
  \end{tabular}
  \caption{Post-training \acceptancelength as a percentage of the matching
  full-KV AL for cross-tokenizer Qwen3.5-4B $\rightarrow$ Kimi-K3 drafting.
  The absolute full-KV AL is $4.834/5.138$ at $T=0/1$.}
  \label{tab:trainable-snapkv-cross-model}
\end{table*}

\definecolor{bsline}{RGB}{35,112,160}

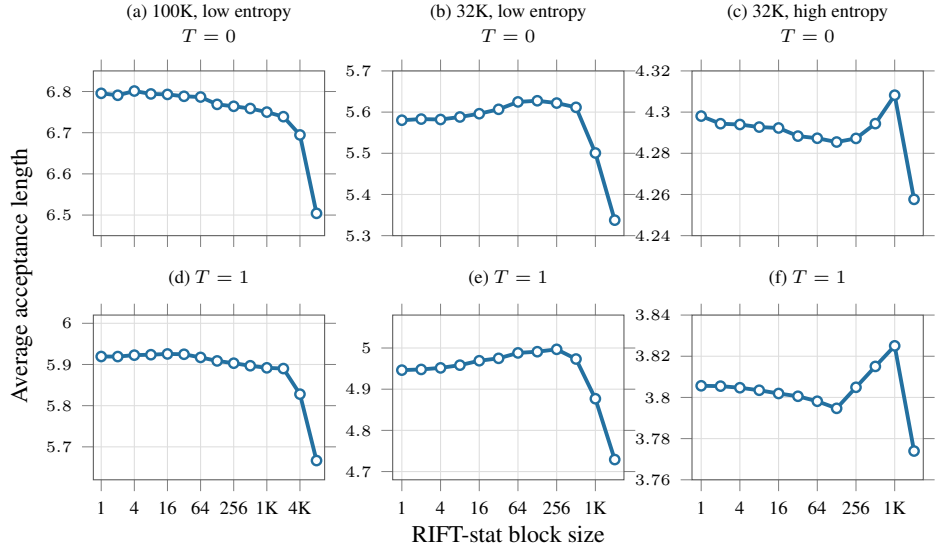
\begin{figure*}[!tbp]
  \centering
  \begin{tikzpicture}
    \pgfplotsset{
      bsaxis/.style={
        width=0.29\textwidth,
        height=0.235\textwidth,
        xmode=log,
        log basis x={2},
        grid=both,
        major grid style={black!13},
        minor grid style={black!6},
        axis line style={black!65},
        tick align=outside,
        xticklabel style={font=\scriptsize},
        yticklabel style={font=\tiny},
        title style={font=\scriptsize, align=center, yshift=1pt},
        scaled ticks=false,
        clip=false,
      },
      bscurve/.style={
        color=bsline,
        line width=1.45pt,
        mark=*,
        mark size=1.75pt,
        mark options={solid,fill=white,line width=0.9pt},
      },
    }
    \begin{groupplot}[
      group style={group size=3 by 2, horizontal sep=0.90cm, vertical sep=1.05cm},
      bsaxis,
    ]
      \nextgroupplot[
        title={\shortstack{(a) 100K, low entropy\\$T=0$}},
        xmin=0.72,xmax=11500,ymin=6.45,ymax=6.85,
        xtick={1,4,16,64,256,1024,4096},
        xticklabels={},
        ytick={6.5,6.6,6.7,6.8},
      ]
        \addplot[bscurve] coordinates {
          (1,6.795596) (2,6.790867) (4,6.801203) (8,6.794122)
          (16,6.792941) (32,6.788413) (64,6.786601) (128,6.768783)
          (256,6.764091) (512,6.758765) (1024,6.750011)
          (2048,6.738902) (4096,6.694594) (8192,6.504162)
        };
      \nextgroupplot[
        title={\shortstack{(b) 32K, low entropy\\$T=0$}},
        xmin=0.72,xmax=2850,ymin=5.30,ymax=5.70,
        xtick={1,4,16,64,256,1024},
        xticklabels={},
        ytick={5.3,5.4,5.5,5.6,5.7},
      ]
        \addplot[bscurve] coordinates {
          (1,5.580298) (2,5.583055) (4,5.582029) (8,5.588139)
          (16,5.596104) (32,5.606782) (64,5.624762) (128,5.627530)
          (256,5.621934) (512,5.611585) (1024,5.500758) (2048,5.337624)
        };
      \nextgroupplot[
        title={\shortstack{(c) 32K, high entropy\\$T=0$}},
        xmin=0.72,xmax=2850,ymin=4.24,ymax=4.32,
        xtick={1,4,16,64,256,1024},
        xticklabels={},
        ytick={4.24,4.26,4.28,4.30,4.32},
      ]
        \addplot[bscurve] coordinates {
          (1,4.298065) (2,4.294365) (4,4.293972) (8,4.292727)
          (16,4.292291) (32,4.288349) (64,4.287308) (128,4.285470)
          (256,4.287224) (512,4.294398) (1024,4.308218) (2048,4.257626)
        };
      \nextgroupplot[
        title={(d) $T=1$},
        xmin=0.72,xmax=11500,ymin=5.62,ymax=6.02,
        xtick={1,4,16,64,256,1024,4096},
        xticklabels={1,4,16,64,256,1K,4K},
        ytick={5.7,5.8,5.9,6.0},
      ]
        \addplot[bscurve] coordinates {
          (1,5.919345) (2,5.919218) (4,5.922653) (8,5.923740)
          (16,5.925613) (32,5.924801) (64,5.917149) (128,5.908641)
          (256,5.903107) (512,5.896934) (1024,5.891869)
          (2048,5.890059) (4096,5.827948) (8192,5.666709)
        };
      \nextgroupplot[
        title={(e) $T=1$},
        xmin=0.72,xmax=2850,ymin=4.68,ymax=5.08,
        xtick={1,4,16,64,256,1024},
        xticklabels={1,4,16,64,256,1K},
        ytick={4.7,4.8,4.9,5.0},
      ]
        \addplot[bscurve] coordinates {
          (1,4.946278) (2,4.947949) (4,4.951719) (8,4.958340)
          (16,4.969087) (32,4.974892) (64,4.988077) (128,4.991027)
          (256,4.996762) (512,4.973264) (1024,4.876869) (2048,4.729136)
        };
      \nextgroupplot[
        title={(f) $T=1$},
        xmin=0.72,xmax=2850,ymin=3.76,ymax=3.84,
        xtick={1,4,16,64,256,1024},
        xticklabels={1,4,16,64,256,1K},
        ytick={3.76,3.78,3.80,3.82,3.84},
      ]
        \addplot[bscurve] coordinates {
          (1,3.805653) (2,3.805486) (4,3.804707) (8,3.803483)
          (16,3.801906) (32,3.800566) (64,3.798146) (128,3.794706)
          (256,3.804924) (512,3.815060) (1024,3.825057) (2048,3.773969)
        };
    \end{groupplot}

    \path (group c1r1.west) -- (group c1r2.west)
      node[midway, xshift=-0.95cm, rotate=90, font=\small]
      {Average acceptance length};
    \path (group c1r2.south) -- (group c3r2.south)
      node[midway, yshift=-0.72cm, font=\small]
      {\riftstat block size};
  \end{tikzpicture}
  \caption{Effect of SnapKV block size for Qwen3.5-4B drafting for
  Qwen3.5-122B with a 50/50 local--remote cache split. The 32K experiments
  use an 8K cache budget; the 100K experiments use a 16K
  budget. The top and bottom rows report $T=0$ and $T=1$,
  respectively. Moderate block sizes are broadly robust, whereas very coarse
  blocks reduce selection resolution and lower \acceptancelength.}
  \label{fig:block-size-ablation}
\end{figure*}

\begin{table*}[!tbp]
    \centering
\footnotesize
    \begin{tabular}{llcc}
        \toprule
        Budget & Selector & $T=0$ & $T=1$ \\
        \midrule
        \multirow{3}{*}{4K} & CriticalKV         & $3.57 \rightarrow 5.02$ $(+40.6\%)$ & $3.15 \rightarrow 4.33$ $(+37.3\%)$ \\
                            & Expected Attention & $3.59 \rightarrow 5.03$ $(+40.2\%)$ & $3.09 \rightarrow 4.30$ $(+39.0\%)$ \\
                            & CurDKV             & $3.29 \rightarrow 4.84$ $(+47.1\%)$ & $2.84 \rightarrow 4.14$ $(+45.5\%)$ \\
        \midrule
        \multirow{3}{*}{8K} & CriticalKV         & $4.13 \rightarrow 5.62$ $(+36.2\%)$ & $3.74 \rightarrow 4.98$
        $(+33.1\%)$ \\
                            & Expected Attention & $4.06 \rightarrow 5.56$ $(+36.9\%)$ & $3.56 \rightarrow 4.83$ $(+35.9\%)$ \\
                            & CurDKV             & $3.76 \rightarrow 5.35$ $(+42.2\%)$ & $3.26 \rightarrow 4.63$ $(+42.1\%)$ \\
        \bottomrule
    \end{tabular}
    \caption{\riftstat extends beyond SnapKV: Qwen3.5-4B draft (Qwen3.5-122B target), 32K ISL, 4K OSL with draft length of 11. Each entry shows baseline AL $\rightarrow$ AL with \riftstat, followed by the relative improvement.}
    \label{tab:static-selectors-sw-ratio-full}
\end{table*}

\section{RIFT-Ada: Additional Results and Analysis}
\label{app:adaptive-snapkv-results}

We provide additional numerical results for \riftada, considering additional datasets, budgets and models. Furthermore, we provide additional information on the calibration and uncertainty of the calibrated threshold.

The example refresh schedule and its score trace are shown in Figure~\ref{fig:adaptive-refresh-monotonic-74}.

\subsection{Additional Results}

\paragraph{Qwen3.5-4B}
Figure~\ref{fig:adaptive-snapkv-results-grid} compares AL--refresh tradeoffs across temperatures, KV budgets, and input lengths using setting-specific controllers, with parameters held fixed across budgets.

\definecolor{gridPeriodic}{RGB}{95,95,95}
\definecolor{gridAdaptive}{RGB}{0,114,178}
\definecolor{gridAlternate}{RGB}{230,159,0}
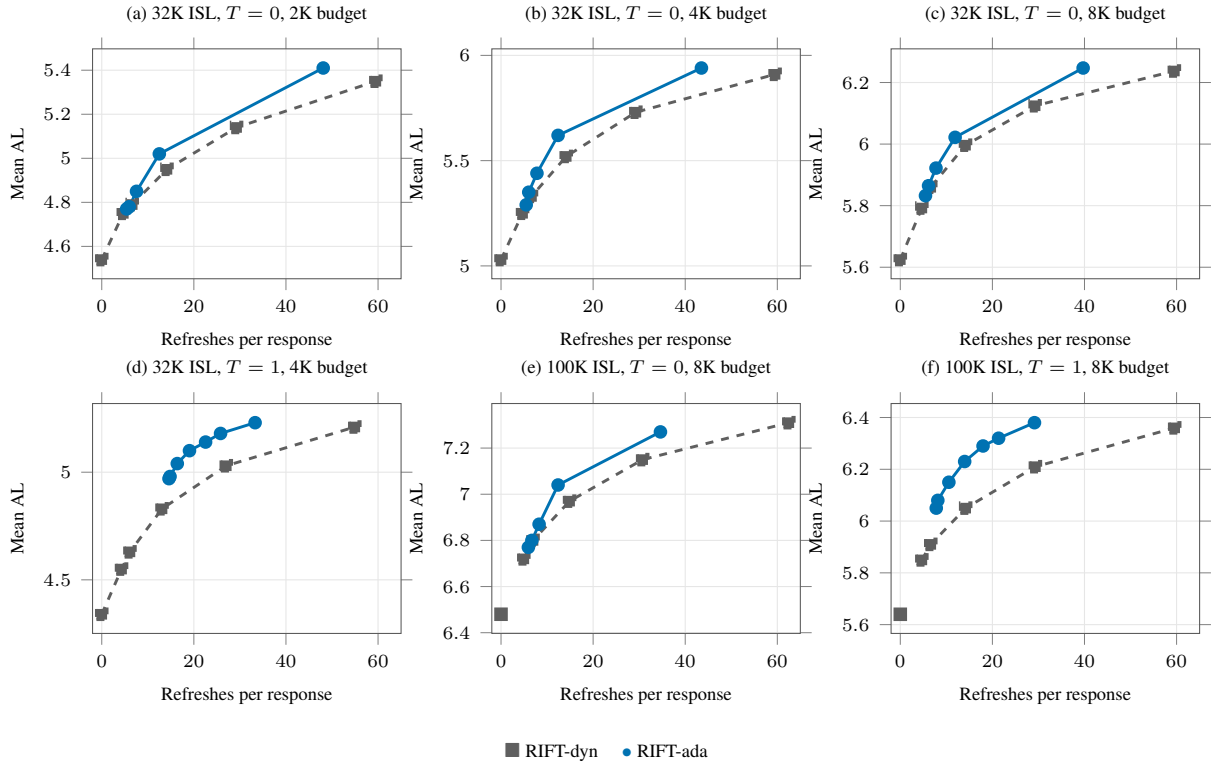
\begin{figure*}[!tbp]
\centering
\begin{tikzpicture}
\begin{groupplot}[
group style={group size=3 by 2,horizontal sep=1.2cm,vertical sep=1.65cm},
scale only axis,width=0.255\linewidth,height=0.19\linewidth,
xmin=-2,xmax=65,xtick={0,20,40,60},xlabel={Refreshes per response},
grid=major,grid style={black!10},axis line style={black!65},tick align=outside,
label style={font=\scriptsize},tick label style={font=\scriptsize},
title style={font=\scriptsize},every axis plot/.append style={line width=1.1pt,mark size=2pt}]
\nextgroupplot[title={(a) 32K ISL, $T=0$, 2K budget},ylabel={Mean AL}]
\addplot[gridPeriodic,dashed,mark=square*] coordinates {(0.00,4.54) (4.68,4.75) (6.61,4.79) (14.16,4.95) (29.28,5.14) (59.52,5.35)};
\addplot[gridAdaptive,mark=*] coordinates {(5.40,4.77) (5.96,4.78) (7.55,4.85) (12.50,5.02) (48.09,5.41)};
\nextgroupplot[title={(b) 32K ISL, $T=0$, 4K budget},ylabel={Mean AL}]
\addplot[gridPeriodic,dashed,mark=square*] coordinates {(0.00,5.03) (4.68,5.25) (6.61,5.33) (14.16,5.52) (29.28,5.73) (59.52,5.91)};
\addplot[gridAdaptive,mark=*] coordinates {(5.47,5.29) (6.02,5.35) (7.80,5.44) (12.39,5.62) (43.52,5.94)};
\nextgroupplot[title={(c) 32K ISL, $T=0$, 8K budget},ylabel={Mean AL}]
\addplot[color=gridPeriodic,dashed,mark=square*] coordinates {
          (0,5.6248) (4.6797,5.7923) (6.6133,5.8597)
          (14.1641,5.9964) (29.2813,6.1245) (59.5156,6.2374)
        };

        \addplot[color=gridAdaptive,mark=*] coordinates {
          (5.4688,5.8324) (6.1406,5.8647) (7.7539,5.9220)
          (11.9063,6.0219) (39.7070,6.2471)
        };

        \nextgroupplot[title={(d) 32K ISL, $T=1$, 4K budget},ylabel={Mean AL}]
\addplot[gridPeriodic,dashed,mark=square*] coordinates {(0.00,4.34) (4.30,4.55) (6.12,4.63) (13.09,4.83) (27.04,5.03) (54.95,5.21)};
\addplot[gridAdaptive,mark=*] coordinates {(14.60,4.97) (14.83,4.98) (16.40,5.04) (19.05,5.10) (22.55,5.14) (25.80,5.18) (33.30,5.23)};
\nextgroupplot[title={(e) 100K ISL, $T=0$, 8K budget},ylabel={Mean AL}]
\addplot[gridPeriodic,only marks,mark=square*,mark size=2pt] coordinates {(0.00,6.48)};
\addplot[gridPeriodic,dashed,mark=square*] coordinates {(4.95,6.72) (6.95,6.80) (14.89,6.97) (30.76,7.15) (62.50,7.31)};
\addplot[gridAdaptive,mark=*] coordinates {(5.94,6.77) (6.70,6.80) (8.27,6.87) (12.39,7.04) (34.62,7.27)};
\nextgroupplot[title={(f) 100K ISL, $T=1$, 8K budget},ylabel={Mean AL}]
\addplot[gridPeriodic,only marks,mark=square*,mark size=2pt] coordinates {(0.00,5.64)};
\addplot[gridPeriodic,dashed,mark=square*] coordinates {(4.68,5.85) (6.62,5.91) (14.20,6.05) (29.34,6.21) (59.59,6.36)};
\addplot[gridAdaptive,mark=*] coordinates {(7.81,6.05) (8.15,6.08) (10.55,6.15) (13.98,6.23) (17.97,6.29) (21.35,6.32) (29.14,6.38)};
\end{groupplot}
\end{tikzpicture}
\par\smallskip
{\scriptsize
\textcolor{gridPeriodic}{$\blacksquare$} \riftdyn \quad
\textcolor{gridAdaptive}{$\bullet$} \riftada
}
\caption{\riftada AL-refresh tradeoffs for Qwen3.5-4B drafting Qwen3.5-122B on SPEED-Bench low-entropy data, with a 4K maximum OSL. Titles specify ISL, temperature, and KV cache budget.}
\label{fig:adaptive-snapkv-results-grid}
\label{fig:adaptive-snapkv-32k-grid}
\label{fig:adaptive-snapkv-100k-grid}
\label{fig:adaptive-snapkv-pareto-temp0}
\end{figure*}

\begingroup
\subsection{Adaptive Refresh for Nemotron Lightning Self-Speculation}
\label{app:nemotron-self-adaptive}
We provide additional analysis of \riftada considering a self-speculative decoding setup with Nemotron3.5-Lightning~\cite{alexiuk2026nemotron35lightning}.
We consider SPEED-Bench 32k low entropy data, with draft length of 11 and 4K OSL.
Following model-specific calibration, we refresh when aggregated acceptance falls below a threshold at $T=0$, or when either acceptance is low or draft entropy is high at $T=1$. Aggregated acceptance uses the last four completed verification steps with exponential smoothing ($\alpha=0.4$). Additionally, we use minimum and maximum refresh ages of 32 and 768 tokens respectively.
As seen in Figure~\ref{fig:nemotron-self-adaptive-pareto}, \riftada significantly improves upon \riftdyn, allowing for similar ALs with a fraction of the overall number of refreshes. These additional results further highlight the necessity of adaptive gating for KV recomputation.

\definecolor{nemoAdaptiveFixed}{RGB}{95,95,95}
\definecolor{nemoAdaptiveCurve}{RGB}{0,114,178}
\definecolor{nemoAdaptivePoint}{RGB}{230,159,0}
\definecolor{nemoAdaptiveGain}{RGB}{204,51,17}
\begin{figure*}[!tbp]
  \centering
  \begin{tikzpicture}
    \begin{groupplot}[
      group style={group size=2 by 1,horizontal sep=1.25cm},
      width=0.44\linewidth,height=0.28\linewidth,
      xlabel={\# Refreshes},xmin=0,xmax=85,
      grid=major,grid style={black!10},axis line style={black!65},
      tick align=outside,label style={font=\small},
      tick label style={font=\scriptsize},title style={font=\small\bfseries},
      every axis plot/.append style={line width=1.15pt,mark size=2.4pt},
    ]
      \nextgroupplot[title={$T=0$},ylabel={AL}]
      \addplot[color=nemoAdaptiveFixed,dashed,mark=square*] coordinates { (0.00000000,6.69756871) (4.55208333,7.33915960) (6.43750000,7.54257857) (13.79166667,7.96398647) (28.48958333,8.32916880) (57.91666667,8.72158187) (116.78125000,9.00539593) (234.48958333,9.18355141) };
      \addplot[color=nemoAdaptiveCurve,mark=*] coordinates { (8.78125000,8.32923933) (13.97916667,8.52389562) (26.47916667,8.71025851) (45.33333333,8.82690329) (74.06250000,8.89571282) };
      \addplot[color=nemoAdaptiveFixed,only marks,mark=o,forget plot] coordinates { (8.78125000,7.67687995) (28.49487247,8.32923933) };
      \draw[<->,nemoAdaptiveGain,line width=0.8pt] (axis cs:8.78125000,7.67687995) -- (axis cs:8.78125000,8.32923933);
      \draw[<->,nemoAdaptiveGain,line width=0.8pt] (axis cs:8.78125000,8.32923933) -- (axis cs:28.49487247,8.32923933);
      \node[anchor=south east,align=right,font=\scriptsize,
        fill=white,fill opacity=0.95,text opacity=1,inner sep=3pt]
        at (rel axis cs:0.98,0.06) {$+0.652$ AL at matched refreshes\\$69.2\%$ fewer at matched AL};
      \nextgroupplot[title={$T=1$},ylabel={}]
      \addplot[color=nemoAdaptiveFixed,dashed,mark=square*] coordinates { (0.00000000,6.93056787) (4.50000000,7.50616079) (6.43750000,7.73561147) (13.76041667,8.09660571) (28.36458333,8.62278317) (57.60416667,9.00707066) (116.08333333,9.30061514) (233.04166667,9.57127725) };
      \addplot[color=nemoAdaptiveCurve,mark=*] coordinates { (7.71875000,8.35880834) (13.06250000,8.66931577) (27.93750000,8.92666269) (49.89583333,9.07376207) (77.52083333,9.19588413) };
      \addplot[color=nemoAdaptiveFixed,only marks,mark=o,forget plot] coordinates { (13.06250000,8.06220086) (31.90514638,8.66931577) };
      \draw[<->,nemoAdaptiveGain,line width=0.8pt] (axis cs:13.06250000,8.06220086) -- (axis cs:13.06250000,8.66931577);
      \draw[<->,nemoAdaptiveGain,line width=0.8pt] (axis cs:13.06250000,8.66931577) -- (axis cs:31.90514638,8.66931577);
      \node[anchor=south east,align=right,font=\scriptsize,
        fill=white,fill opacity=0.95,text opacity=1,inner sep=3pt]
        at (rel axis cs:0.98,0.06) {$+0.607$ AL at matched refreshes\\$59.1\%$ fewer at matched AL};
    \end{groupplot}
  \end{tikzpicture}
  \par\smallskip
  {\scriptsize
    \textcolor{nemoAdaptiveFixed}{$\blacksquare$} \riftdyn \quad
    \textcolor{nemoAdaptiveCurve}{$\bullet$} \riftada}

  \caption{Nemotron3.5-Lightning self-speculation \riftada refresh - AL tradeoff. SPEED-Bench 32k ISL, low entropy category.}
  \label{fig:nemotron-self-adaptive-pareto}
\end{figure*}
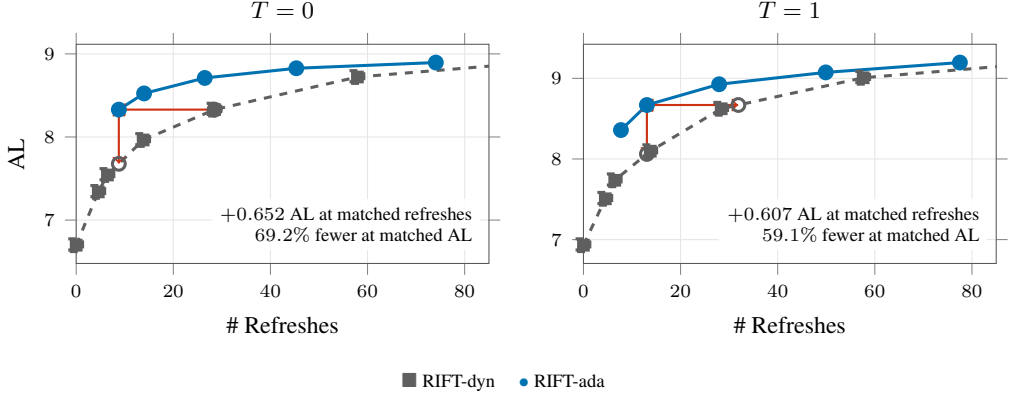
\endgroup

\subsection{Adaptive-Refresh Policies and Calibration}
\label{app:adaptive-refresh-audit}

\textbf{Calibration and policy selection.}
We organize controller development around three sets:
calibration, policy selection, and confirmation.
We begin by collecting the signals on the calibration set, from which we determine the threshold search sets. The set of thresholds is determined from the corresponding percentiles which are derived from the signals' statistics.
Next, the policy-selection set compares measured AL-refresh tradeoffs for the various threshold candidates. We calibrate the controller separately for each draft/target-model and temperature setting.
For the reported budget sweeps, controller parameters
are kept fixed across budgets.

\textbf{Selected policies.}
Generation settings change draft entropy, acceptance
statistics, and the useful refresh operating range,
motivating separate controller calibration.
The selected policies therefore differ across models
and temperatures, and may be further tailored to a
particular generation setting.
Acceptance-based rules use exponential weighting of 4 completed verification steps with coefficient $0.4$.
Entropy is averaged over the latest decision stride.
Signals are updated after each completed decision stride, before deciding whether to refresh. Query drift compares the current grouped query summary with the summary stored at the last refresh.

The \riftada curve in Figure~\ref{fig:adaptive-snapkv-pareto} uses an \texttt{or}-combined rule.
Let $A_t$ be the four-completed-step exponentially weighted acceptance length ($\alpha=0.4$), $H_t$ the mean draft entropy over the latest decision stride, and $D^q_t$ the mean query drift. Let $A_{\mathrm{ref}}$ and $H_{\mathrm{ref}}$ denote the post-refresh references. We denote the collection of signals with $z_t$. Thus,  the signal trigger is given by
\begin{equation*}
\begin{aligned}
\pi(z_t) ={}& (H_t \ge 2.11)\\
 &\mathbin{\texttt{or}}\ (A_t \le 2.19)\\
 &\mathbin{\texttt{or}}\ (H_t-H_{\mathrm{ref}} \ge 1.35)\\
 &\mathbin{\texttt{or}}\ (A_{\mathrm{ref}}-A_t \ge 4.84)\\
 &\mathbin{\texttt{or}}\ (D^q_t \ge \tau_D).
\end{aligned}
\end{equation*}
To obtain the curve, we keep the other thresholds fixed and sweep $\tau_D$ over $\{0.5,0.43,0.33,0.26,0.22,0.2,0.16\}$.

\paragraph{Uncertainty and evaluation scope.}
We calibrate separate controllers for each temperature, using disjoint calibration, evaluation, and confirmation subsets during development. We compare fitted monotonic-score policies with acceptance/entropy level and deterioration rules, including \texttt{and}/\texttt{or} combinations with query drift.
For the Qwen results in Figures~\ref{fig:adaptive-snapkv-results-grid}(c) ($T=0$) and~\ref{fig:adaptive-snapkv-pareto} ($T=1$), the selected policies use level and deterioration signals at $T=0$, and an additional query-drift \texttt{or} trigger at $T=1$, reducing refreshes at matched AL by $53.5\%$ and $51.1\%$, respectively. Paired-bootstrap 95\% confidence intervals exclude zero for both improvements over periodic refresh, conditional on the selected policies. The final results include the development subsets, so these results describe selected-policy performance rather than an untouched held-out evaluation.

\paragraph{Measuring improvement in the refresh-AL tradeoff.}
For an adaptive operating point $(R,A)$, where $R$ is the average number of refreshes and $A$ is AL, we report
\begin{equation}
\begin{aligned}
\Delta A\vert_R &= A-A_{\mathrm{fix}}(R),\\
\Delta R\vert_A &=100\!\left(1-\frac{R}{R_{\mathrm{fix}}(A)}\right)\%.
\end{aligned}
    \label{eq:adaptive-fixed-axis-metrics}
\end{equation}
Here $A_{\mathrm{fix}}(R)$ is periodic refresh's AL at matched refresh count, and $R_{\mathrm{fix}}(A)$ is its refresh count at matched AL.

\section{Beyond Speculative Decoding: Long-Context Quality Evaluation}
\label{app:long-context-quality-protocol}
We provide preliminary evidence for \riftstat's performance beyond speculative decoding in a long-OSL setting. We evaluate Qwen3.5-4B on LongBench-v1 and LongBench-v2~\cite{bai2024longbench,bai2025longbench}.

\textbf{Evaluation protocol.}
To meet a long OSL setup, we enforce a 4K reasoning phase. Following prefill, the model greedily generates up to 4,096 reasoning tokens. Then, it receives a task-specific final-answer request. Generation continues with the same cache, without re-prefill or reselection. Answer limits are 32-512 tokens for LongBench-v1 and 128 for LongBench-v2. We report the average of task scores. We count parse failures as incorrect (e.g., no valid choice in a multi-option question).
We compare SnapKV (with a $64$ token SW) and \riftstat.

\textbf{Results.}
As seen in Table~\ref{tab:sw-ratio-beyond-specdec} \riftstat uniformly outperforms SnapKV at every tested budget. At 8K, it retains 96.4\% and 89.9\% of full-KV scores on v1 and v2, respectively. These results suggest that recency allocation can also benefit standalone generation. We note that, while promising, these are preliminary results, and a comprehensive evaluation beyond speculative decoding is left to future work.

\begin{table*}[!tbp]
    \centering
\footnotesize

    \begin{tabular}{llrr}
        \toprule
        Benchmark & Budget (Avg. compression) & SnapKV & \riftstat  \\
        \midrule
        \multirow{4}{*}{LongBench-v1}
          & 1K (9.95$\times$) & 60.7 & 76.2 (+15.5) \\
          & 2K (4.98$\times$) & 61.7 & 81.8 (+20.1) \\
          & 4K (2.49$\times$) & 63.4 & 89.5 (+26.1) \\
          & 8K (1.24$\times$) & 67.2 & 96.4 (+29.2) \\
        \midrule
        \multirow{4}{*}{LongBench-v2}
          & 1K (119.41$\times$) & 60.3 & 61.9 (+1.6) \\
          & 2K (59.70$\times$) & 67.7 & 70.4 (+2.6) \\
          & 4K (29.85$\times$) & 68.3 & 75.7 (+7.4) \\
          & 8K (14.93$\times$) & 69.3 & 89.9 (+20.6) \\
        \bottomrule
    \end{tabular}

    \caption{Full-KV score retention (\%) for Qwen3.5-4B under the modified LongBench protocol described above. Absolute full-KV scores are $47.607$ for LongBench-v1 and $38.569$ for LongBench-v2.}
    \label{tab:sw-ratio-beyond-specdec}
\end{table*}

\begingroup
\section{\rift Performance: Task Breakdown}
\label{app:domains-breakdown}
To provide an additional point of view into \rift's performance, we break down acceptance length by SPEED-Bench task. We consider Qwen3.5-4B drafting for Qwen3.5-122B and consider data from ISL splits 16K, 32K and 100K. Table~\ref{tab:domains-breakdown} presents nine selected categories. It can be seen that the order of performance (SnapKV<SW<\riftstat<\riftdyn) is maintained across all categories, across various generation modes induces by different tasks
\par
\endgroup

\begin{table*}[tbp]
\centering
\small
\setlength{\tabcolsep}{4pt}
\begin{tabular}{lrrrrr}
\toprule
Task & SnapKV & SWA & \riftstat & \riftdyn & \riftada \\
\midrule
\multicolumn{6}{c}{$T=0$} \\
\midrule
Sorting & 3.65 & 3.96 & 4.84 & 5.35 & 5.61 \\
Creative writing & 3.90 & 4.86 & 5.79 & 6.29 & 6.28 \\
NIAH & 4.30 & 4.86 & 5.50 & 6.16 & 6.18 \\
Code debugging & 3.76 & 5.02 & 5.58 & 6.67 & 6.80 \\
Creative writing & 3.03 & 3.91 & 4.18 & 4.39 & 4.42 \\
Domain QA --- Math & 3.26 & 5.64 & 5.82 & 5.91 & 5.91 \\
Domain QA --- Chemistry & 3.75 & 6.22 & 6.45 & 6.47 & 6.50 \\
Domain QA --- Biology/Medicine & 3.50 & 5.78 & 5.98 & 6.03 & 6.03 \\
Domain QA --- Engineering & 4.39 & 6.25 & 6.44 & 6.55 & 6.58 \\
\midrule
\multicolumn{6}{c}{$T=1$} \\
\midrule
Sorting & 3.18 & 3.57 & 4.22 & 4.58 & 4.78 \\
Creative writing & 3.13 & 4.14 & 4.85 & 5.40 & 5.39 \\
NIAH & 3.66 & 4.22 & 4.72 & 5.35 & 5.32 \\
Code debugging & 3.25 & 4.52 & 4.86 & 5.77 & 6.07 \\
Creative writing & 2.76 & 3.55 & 3.76 & 3.90 & 3.93 \\
Domain QA --- Math & 2.83 & 5.19 & 5.32 & 5.39 & 5.39 \\
Domain QA --- Chemistry & 3.26 & 5.84 & 5.95 & 6.01 & 6.01 \\
Domain QA --- Biology/Medicine & 3.15 & 5.34 & 5.45 & 5.52 & 5.53 \\
Domain QA --- Engineering & 3.79 & 5.90 & 6.00 & 6.09 & 6.10 \\
\bottomrule
\end{tabular}
\caption{Mean AL using SPEED-Bench data of ISLs 16K, 32K and 100K, at $T=0$ and $T=1$. Draft model is Qwen3.5-4B drafting for Qwen3.5-122B. Draft length is 11 and max OSL is 4K.}
\label{tab:domains-breakdown}
\end{table*}

\begingroup
\section{Tokenizer Adaptation for Cross-Family Drafting}
\label{app:tokenizer-adaptation}

We adopt the following tokenizer-adaptation procedure to initialize a cross-family draft before further draft--target alignment:
\begin{enumerate}[leftmargin=*,nosep]
    \item \textbf{Resize.} Resize draft input embeddings and the LM-head to the target's configured vocabulary size, retaining the draft's hidden dimension.
    \item \textbf{Copy matching tokens.} For each target token, first try a manually role-matched control token (for special tokens). If the token is no a special token, look for an exact match, and if such a match is not found look for a normalized match in the draft vocabulary.
    \item \textbf{Initialize remaining tokens.} Retokenize unmatched text with the draft tokenizer and combine up to four embeddings using geometrically decreasing weights.
    \item \textbf{Align outputs.} Apply the same mapping to an untied output head, preserve tied embeddings, and adopt the target tokenizer's beginning-of-sequence, end-of-sequence, and padding IDs.
\end{enumerate}
Once the draft's vocabulary and corresponding embedding matrices are aligned, we perform a knowledge distillation scheme over the Nemotron post-training V2 dataset over 10K steps.
\endgroup

\section{Comparison With Additional KV Compression Policies}
\label{app:decode-policy-comparison}
We provide additional comparisons with KV compression methods, complementing the results in the main paper. We evaluate on SPEED-Bench with Qwen3.5-4B drafting for Qwen3.5-122B and, in Table~\ref{tab:attention-sinks-comparison}, Llama3.1-8B drafting for Llama3.1-70B. We consider 4K OSL and a draft length of 11.
We consider the following methods:
\begin{enumerate}
    \item StreamingLLM~\cite{xiao2024streamingllm}, which retains a SW of recent tokens together with four initial attention-sink tokens.
    \item H$_2$O~\cite{zhang2023h2o}, which updates accumulated attention scores at each decoding step and retains a combination of recent tokens and heavy hitters, permanently evicting other positions.
    \item TriAttention~\cite{mao2026triattention}, which uses calibrated pre-RoPE query statistics to estimate key importance through RoPE-induced trigonometric distance preferences and norm-based scoring. Specifically, we estimate the mean pre-RoPE query vectors from full-prefill statistics and hold them fixed during decoding, and consider a refresh interval similar to \riftdyn of 128 tokens.
\end{enumerate}
Tables~\ref{tab:attention-sinks-comparison} and~\ref{tab:decode-time-comparison} show higher AL for \rift than StreamingLLM across the reported settings.
For the other decode-time baselines, \riftdyn and \riftada outperform both H$_2$O and TriAttention in every reported setting, while H$_2$O generally outperforms \riftstat (Table~\ref{tab:decode-time-comparison}).

\begin{table*}[!t]
  \centering
  \footnotesize
  \begin{tabular}{llcrrrr}
    \toprule
    {Model} & {ISL} & {$T$} & {StreamingLLM} & \riftstat & \riftdyn & {\riftada} \\
    \midrule
    \multirow{4}{*}{{Qwen3.5}} & \multirow{2}{*}{{32K}} & {0} & {5.34} & {5.58} & {6.08} & {6.17} \\
     &  & {1} & {4.73} & {4.92} & {5.40} & {5.49} \\
    \cmidrule(lr){2-7}
     & \multirow{2}{*}{{100K}} & {0} & {5.60} & {5.77} & {6.61} & {6.73} \\
     &  & {1} & {4.98} & {5.07} & {5.76} & {5.92} \\
    \midrule
    \multirow{4}{*}{{Llama3.1}} & \multirow{2}{*}{{32K}} & {0} & {5.87} & {5.87} & {6.58} & {7.08} \\
     &  & {1} & {4.20} & {4.41} & {4.89} & {5.41} \\
    \cmidrule(lr){2-7}
     & \multirow{2}{*}{{100K}} & {0} & {4.82} & {5.32} & {5.65} & {6.23} \\
     &  & {1} & {3.35} & {3.58} & {3.70} & {3.84} \\
    \bottomrule
  \end{tabular}
  \caption{Mean AL on SPEED-Bench low-entropy data. Qwen3.5-4B drafts for Qwen3.5-122B and Llama3.1-8B for Llama3.1-70B. All rows use an 8K budget, draft length 11, and a 4K maximum OSL. StreamingLLM retains an 8K window plus four sinks.}
  \label{tab:attention-sinks-comparison}
\end{table*}

\begin{table*}[!t]
  \centering
\footnotesize
  \small
  \begin{tabular}{crrrrrrr}
    \toprule
    $T$ & Budget & StreamingLLM & H$_2$O & TriAttention & \riftstat & \riftdyn & \riftada \\
    \midrule
    \multicolumn{8}{c}{\textbf{ISL 32K}} \\
    \midrule
0 & 2K & {4.52} & 4.68 & 4.54 & 4.54 & 5.14 & 5.28 \\
     & 4K & {4.79} & 5.25 & 4.97 & 5.03 & 5.73 & 5.82 \\
     & 8K & {5.34} & 5.73 & 5.43 & 5.62 & 6.12 & 6.17 \\
    \midrule
    1 & 2K & {3.90} & 4.06 & 3.86 & 3.94 & 4.47 & 4.54 \\
     & 4K & {4.17} & 4.53 & 4.27 & 4.34 & 5.03 & 5.10 \\
     & 8K & {4.73} & 5.03 & 4.77 & 4.99 & 5.48 & 5.51 \\
    \midrule
    \multicolumn{8}{c}{\textbf{ISL 100K}} \\
    \midrule
0 & 2K & {4.84} & 5.35 & 5.17 & 5.39 & 6.11 & 6.33 \\
     & 4K & {5.30} & 5.92 & 5.64 & 5.81 & 6.75 & 6.91 \\
     & 8K & {5.60} & 6.65 & 6.10 & 6.40 & 7.14 & 7.17 \\
    \midrule
    1 & 2K & {4.28} & 4.47 & 4.25 & 4.54 & 5.16 & 5.46 \\
     & 4K & {4.67} & 5.06 & 4.69 & 4.98 & 5.80 & 6.09 \\
     & 8K & {4.98} & 5.77 & 5.11 & 5.55 & 6.21 & 6.36 \\
    \bottomrule
  \end{tabular}
  \caption{Mean AL on SPEED-Bench low entropy for Qwen3.5-4B drafting Qwen3.5-122B, with draft length 11 and a 4K maximum OSL. StreamingLLM uses the stated window plus four sinks.
  }
  \label{tab:decode-time-comparison}
\end{table*}

\section{Draft Compressibility Training}
\label{app:draft-compressibility-results}

\subsection{Mask Construction}
To generate a SnapKV attention mask, we begin with a frozen dense forward pass and capture the post-RoPE queries and keys in every full-attention layer. Let $\mathcal{O}$ denote the observation window's queries and $\mathcal{G}(h)$ the query heads that share KV head $h$. We denote the dense post-softmax attention weight from query $q$ in head $g$ to key position $k$ in KV head $h$ by $A_{g,h}(q,k)$. The importance of key position $k$ is
\begin{equation}
    a_{h}(k)
    = \frac{1}{|\mathcal{G}(h)|}
      \sum_{g\in\mathcal{G}(h)}\sum_{q\in\mathcal{O}}
      A_{g,h}(q,k).
    \label{eq:trainable-snapkv-score}
\end{equation}

\subsection{Training Configuration}
\label{app:training-configuration}
We consider $10$K training steps using teacher-generated rollouts from the Nemotron V2 post-training dataset, using a batch size of $64$. The teacher responses and logits were generated at $T=1$.
We train with a categorical distribution over
$p\in\{0.1,0.3,0.5,0.7,0.9\}$ whose values are $\{0.05,0.10,0.20,0.25,0.40\}$ respectively. We use a SW ratio $\rho=0.25$ during training.
We consider KD temperature $T_{\mathrm{KD}}=2$, and $(\lambda_{\mathrm{CE}},\lambda_{\mathrm{KD}})=(0.25,0.75)$.
In evaluation, we consider an OSL of $4$K and a draft length of 11, drafting on SPEED-Bench's 32k throughput split, low entropy category.
Additional results and ablation studies follow below.

\subsection{Ablation Studies}
\label{app:draft-compressibility-ablations}
We present ablation studies on various components in the draft compressibility training scheme, motivating the design choices in this work. Unless stated otherwise, we train Qwen3.5-4B drafting for Qwen3.5-122B and evaluate ALs on SPEED-Bench 32K ISL split, low entropy category.

\textbf{Masking Strategy}
We test the gain of the attention-score-based mask by comparing it with similar training runs that use different masking strategies, such as random or SWA-based masks.
As seen in Table~\ref{tab:qwen35_masking_ablation_snapkv}, the attention-based masking scheme outperforms every random and sliding-window control at temperatures 0 and 1 and at all three considered KV budgets. It is worth noting that using an attention-based mask induces additional complexity to the overall training scheme, as we need to perform an additional dense forward pass.

\textbf{Attention mask composition}
We study the effect of the distribution of the sparsity parameter, which determines the ratio of attention to mask out during training. Specifically, we consider dynamic $p$ values, in which $p$ is sampled from a given categorical distribution on a subset of $[0,1]$. All distributions sample $p\in\{0.1,0.3,0.5,0.7,0.9\}$. High biased samples with probabilities $\{0.05,0.10,0.20,0.25,0.40\}$, ``soft'' and ``sharp'' use probabilities $\{0.10,0.15,0.20,0.25,0.30\}$ and $\{0.03,0.07,0.15,0.25,0.50\}$, respectively. Additionally, we test the effect of the training-time SW ratio value $\rho$.
As seen in Table~\ref{tab:qwen35_training_mask_hyperparameters}, the high-biased $p$ distribution attains the best results, while the value of $\rho$ does not have a significant effect on performance.

\textbf{Optimization and Parameterization}
We examine the effect of the fine-tune hyperparameters on the training scheme. Specifically, we are examine the effect of the number of trainable parameters on the resulting model's performance and the learning rate behavior. Table~\ref{tab:qwen35_ft_lora_comparison}, compares LoRA rank and learning-rate schedule for the same Qwen pair, averaging over 2K, 4K, and 8K budgets.
As expected, full model fine tune results in slightly higher ALs than LoRA. However, the computational overhead of a full FT is significant. To that end, when prioritizing performance and the resulting task does not involve self-speculation, we suggest full-FT, while LoRA is preferable when we are interested in a lightweight modification.

\subsection{Additional Draft Compressibility Training Results}
\label{app:draft-compressibility-additional-results}
We present additional results on draft compressibility, demonstrating its effect in additional settings.

\textbf{Additional evaluations on long context}
Table~\ref{tab:trainable-snapkv-qwen4b-100k} extends the Qwen3.5-4B draft (Qwen3.5-122B target) training result to long contexts by evaluating on SPEED-Bench 100k ISL split.
It can be seen that the gains are also evident in longer context drafting, with higher gains in non-greedy decoding settings.

\textbf{Effect of compressibility training to SnapKV.}
We demonstrate the effect of the proposed draft compressibility training on the baseline SnapKV method, considering Qwen3.5-4B drafting for Qwen3.5-122B.
As seen in Table~\ref{tab:trainable-snapkv-qwen4b-r64-100k}, SnapKV benefits dramatically from the proposed compressibility training scheme. However, we attribute the magnitude of that gain to the initial method's AL, which was significantly lower than \riftstat (As reported in the main paper).

\textbf{Additional draft training}
We present the effect of draft compressibility training on additional draft models.
Table~\ref{tab:trainable-snapkv-qwen08b-improvement} presents the effect on Qwen3.5-0.8B (drafting for Qwen3.5-122B).  Gains are modest at $T=0$ but reach up to $~10\%$ at $T=1$.  The corresponding untrained policy comparison appears in Table~\ref{tab:trainable-snapkv-qwen08b-base-relative}.
Next, we consider a larger draft model. Specifically, we train Nemotron3.5-Lightning with a Nemotron-3-Ultra teacher.
The training recipe follows the one described in Section~\ref{sec:draft-compressibility-training}, with a single slight change: Following a empirical sweep, we replace the KL loss term with a skewed KL loss, which replaces the teacher's logits with a linear interpolation of the student and teacher's logits. The underlying potential reason for the performance improvement of skewed KL is that skewed KL compensates for large model distribution discrepancies, which may follow due to the large parameter count ratio between the draft and the target. As shown in Table~\ref{tab:nemotron35-lightning-ultra-training}, compressibility training improves Lightning's AL by up to approximately $16.7\%$ (3.72 to 4.34 for \riftdyn at 2K and $T=1$). At 8K and $T=0$, its trained \riftdyn AL of 5.27 corresponds to 97.8\% of the original draft's full-KV AL (5.39), or 91.7\% of the trained draft's full-KV AL (5.75).

\subsection{Additional \riftada Results on Draft Compressibility}
We provide additional results on the effect on draft compressibility training on \riftada.
We consider Qwen3.5-4B and Qwen3.5-0.8B, drafting for Qwen3.5-122B.
As seen in  Table~\ref{tab:trainable-snapkv-adaptive-improvement} training improves AL in every reported setting.

\begin{table*}[!tbp]
\centering
\footnotesize
\small
\setlength{\tabcolsep}{5pt}
\begin{tabular}{lcccccc}
\toprule
& \multicolumn{3}{c}{$T=0$}
& \multicolumn{3}{c}{$T=1$} \\
\cmidrule(lr){2-4}
\cmidrule(lr){5-7}
Training configuration
& 2K & 4K & 8K
& 2K & 4K & 8K \\
\midrule
Random, dynamic $p$
& 4.67 & 5.15 & 5.70
& 4.32 & 4.71 & 5.34 \\

Random, $p=0.1$
& 4.31 & 4.76 & 5.20
& 3.99 & 4.37 & 4.87 \\

Random, $p=0.3$
& 4.56 & 5.04 & 5.56
& 4.22 & 4.62 & 5.21 \\

Random, $p=0.5$
& 4.59 & 5.09 & 5.67
& 4.24 & 4.66 & 5.29 \\

Random, $p=0.7$
& 4.63 & 5.12 & 5.70
& 4.25 & 4.66 & 5.31 \\

Random, $p=0.9$
& 4.62 & 5.12 & 5.69
& 4.23 & 4.64 & 5.29 \\

SWA, $B=64$
& 4.22 & 4.47 & 4.74
& 3.94 & 4.14 & 4.47 \\

SWA, $B=128$
& 4.36 & 4.63 & 4.93
& 4.07 & 4.29 & 4.65 \\

SWA, $B=256$
& 4.49 & 4.78 & 5.10
& 4.18 & 4.43 & 4.81 \\

SWA, $B=512$
& 4.64 & 4.97 & 5.36
& 4.32 & 4.60 & 5.04 \\
\midrule
\textbf{Attention-based mask}
& \textbf{4.89} & \textbf{5.37} & \textbf{5.94}
& \textbf{4.58} & \textbf{4.98} & \textbf{5.60} \\
\bottomrule
\end{tabular}
\caption{
AL of the trained draft model (Qwen3.5-4B) under various masking strategies. 'dynamic $p$' refers to the same $p$ allocation used in the main draft-compressibility results.
}
\label{tab:qwen35_masking_ablation_snapkv}
\end{table*}

\begin{table*}[!tbp]
\centering
\footnotesize
\small
\setlength{\tabcolsep}{5pt}
\begin{tabular}{lcccccc}
\toprule
& \multicolumn{3}{c}{\riftstat}
& \multicolumn{3}{c}{\riftdyn} \\
\cmidrule(lr){2-4}
\cmidrule(lr){5-7}
Training configuration
& 2K & 4K & 8K
& 2K & 4K & 8K \\
\midrule
\multicolumn{7}{l}{\textit{(a) Sliding-window ratio $\rho$ during training}} \\
$\rho=0.10$ & 4.67 & 5.19 & 5.80 & 5.36 & 5.92 & 6.28 \\
$\rho=0.25$ & 4.69 & 5.21 & 5.82 & 5.39 & 5.92 & 6.29 \\
$\rho=0.50$ & 4.72 & 5.22 & 5.81 & 5.40 & 5.93 & 6.29 \\
$\rho=0.75$ & 4.72 & 5.21 & 5.79 & 5.38 & 5.93 & 6.30 \\
$\rho=0.90$ & 4.72 & 5.20 & 5.78 & 5.35 & 5.88 & 6.28 \\
\midrule
\multicolumn{7}{l}{\textit{(b) Sampling distribution over $p_{\mathrm{keep}}$}} \\
Soft high-biased  & 4.73 & 5.21 & 5.81 & 5.41 & 5.92 & 6.28 \\
High-biased       & 4.72 & 5.22 & 5.81 & 5.40 & 5.93 & 6.29 \\
Sharp high-biased & 4.70 & 5.19 & 5.80 & 5.38 & 5.93 & 6.32 \\
\bottomrule
\end{tabular}
\caption{
Training-mask hyperparameter ablations for Qwen3.5-4B drafting
Qwen3.5-122B on SPEED-Bench 32K low-entropy.
All checkpoints are evaluated with
$\rho=0.5$ with $T=0$.
}
\label{tab:qwen35_training_mask_hyperparameters}
\end{table*}

\begin{table*}[!tbp]
\centering
\footnotesize
\small
\setlength{\tabcolsep}{5pt}
\begin{tabular}{lcccc}
\toprule
& \multicolumn{2}{c}{$T=0$}
& \multicolumn{2}{c}{$T=1$} \\
\cmidrule(lr){2-3}
\cmidrule(lr){4-5}
Training configuration
& \riftstat & \riftdyn
& \riftstat & \riftdyn \\
\midrule
LoRA-32, constant $10^{-5}$
& $\mathbf{5.30 \pm 0.13}$
& $\mathbf{5.85 \pm 0.13}$
& $4.93 \pm 0.11$
& $\mathbf{5.46 \pm 0.12}$ \\

LoRA-128, decay from $10^{-5}$
& $5.29 \pm 0.13$
& $\mathbf{5.85 \pm 0.13}$
& $4.93 \pm 0.12$
& $5.45 \pm 0.13$ \\

Full FT, constant $5{\times}10^{-6}$
& $5.29 \pm 0.13$
& $5.82 \pm 0.13$
& $4.91 \pm 0.12$
& $5.43 \pm 0.13$ \\

Full FT, decay from $10^{-5}$
& $\mathbf{5.30 \pm 0.13}$
& $\mathbf{5.85 \pm 0.13}$
& $\mathbf{4.94 \pm 0.12}$
& $\mathbf{5.46 \pm 0.13}$ \\
\bottomrule
\end{tabular}
\caption{
Training hyperparameter comparison. Results are reported in average ALs with 95\% confidence intervals.
}
\label{tab:qwen35_ft_lora_comparison}
\end{table*}

\begin{table*}[!tbp]
  \centering
\footnotesize
  \setlength{\tabcolsep}{4pt}

  \begin{tabular}{c@{\hspace{8pt}}ccc@{\hspace{8pt}}ccc@{\hspace{8pt}}ccc}
    \toprule
    & \multicolumn{3}{c}{\riftstat}
    & \multicolumn{3}{c}{\riftdyn}
    & \multicolumn{3}{c}{\riftada} \\
    \cmidrule(lr){2-4}\cmidrule(lr){5-7}\cmidrule(lr){8-10}
    $T$ & 2K & 4K & 8K & 2K & 4K & 8K & 2K & 4K & 8K \\
    \midrule
    0 & $+4.05\%$ & $+2.24\%$ & $+0.97\%$
      & $+4.90\%$ & $+2.54\%$ & $+0.97\%$
      & $+2.57\%$ & $+0.37\%$ & $-0.34\%$ \\
    1 & $+12.50\%$ & $+9.87\%$ & $+7.03\%$
      & $+12.04\%$ & $+7.95\%$ & $+5.12\%$
      & $+9.80\%$ & $+5.49\%$ & $+3.79\%$ \\
    \bottomrule
  \end{tabular}
  \caption{Relative AL improvement from compressibility training for
  Qwen3.5-4B drafting Qwen3.5-122B on SPEED-Bench 100K. 
  }
  \label{tab:trainable-snapkv-qwen4b-100k}
\end{table*}

\begin{table*}[!tbp]
  \centering
\footnotesize
  \setlength{\tabcolsep}{8pt}
  \begin{tabular}{cccc}
    \toprule
    $T$ & 2K & 4K & 8K \\
    \midrule
    0 & $+15.57\%$ & $+13.50\%$ & $+11.52\%$ \\
    1 & $+26.54\%$ & $+23.80\%$ & $+21.07\%$ \\
    \bottomrule
  \end{tabular}
  \caption{Relative AL improvement from draft compressibility training for SnapKV:
  Qwen3.5-4B draft (Qwen3.5-122B target), 100K low-entropy split.
  }
  \label{tab:trainable-snapkv-qwen4b-r64-100k}
\end{table*}

\begin{table*}[!tbp]
  \centering
\footnotesize
  \setlength{\tabcolsep}{5pt}
  \begin{tabular}{llccrrrr}
    \toprule
    ISL & Entropy & $T$
      & Full KV & SW & \riftstat & \riftdyn \\
    \midrule
    \multirow{4}{*}{32K}
      & Low  & 0 & $+0.55\%$ & $+0.90\%$ & $+0.84\%$ & $+0.81\%$ \\
      & Low  & 1  & $+7.77\%$ & $+6.43\%$ & $+9.71\%$ & $+8.87\%$ \\
      & High & 0  & $+0.82\%$ & $+0.80\%$ & $+0.98\%$ & $+0.84\%$ \\
      & High & 1  & $+8.95\%$ & $+7.30\%$ & $+9.33\%$ & $+9.32\%$ \\
    \midrule
    \multirow{3}{*}{100K}
      & Low  & 0  & $+0.00\%$ & $+1.73\%$ & $+0.90\%$ & $+0.77\%$ \\
      & Low  & 1  & $+5.90\%$ & $+5.70\%$ & $+8.24\%$ & $+7.68\%$ \\
      & High & 0  & $+0.65\%$ & $-0.25\%$ & $+0.71\%$ & $+0.52\%$ \\
    \bottomrule
  \end{tabular}
  \caption{Relative AL improvement from training the Qwen3.5-0.8B draft, measured on SPEED-Bench at an 8K cache budget against the corresponding untrained draft.
  }
  \label{tab:trainable-snapkv-qwen08b-improvement}
\end{table*}

\begin{table*}[!tbp]
\centering
\footnotesize
\small
\setlength{\tabcolsep}{5pt}
\begin{tabular}{llrrrr}
\toprule
& & \multicolumn{2}{c}{$T=0$}
& \multicolumn{2}{c}{$T=1$} \\
\cmidrule(lr){3-4}
\cmidrule(lr){5-6}
Inference method & Budget & Base & Trained & Base & Trained \\
\midrule
Full KV & -- & 5.39 & 5.75 & 4.86 & 5.49 \\
\midrule
\multirow{3}{*}{Sliding window}
& 2K & 3.62 & 3.87 & 3.21 & 3.73 \\
& 4K & 3.92 & 4.19 & 3.49 & 4.02 \\
& 8K & 4.31 & 4.60 & 3.86 & 4.43 \\
\midrule
\multirow{3}{*}{\riftstat}
& 2K & 3.75 & 4.04 & 3.39 & 3.92 \\
& 4K & 3.99 & 4.29 & 3.60 & 4.16 \\
& 8K & 4.35 & 4.68 & 3.93 & 4.52 \\
\midrule
\multirow{3}{*}{\riftdyn}
& 2K & 4.16 & 4.49 & 3.72 & 4.34 \\
& 4K & 4.57 & 4.93 & 4.08 & 4.72 \\
& 8K & 4.99 & 5.27 & 4.45 & 5.04 \\
\bottomrule
\end{tabular}
\caption{
Average accepted length before and after compressibility training of
Nemotron3.5-Lightning drafting Nemotron-3-Ultra on SPEED-Bench 32K
low-entropy. The trained checkpoint uses skewed KL with
$\alpha=0.5$. Static and Dynamic SnapKV use a sliding-window ratio of
$\rho=0.5$.
}
\label{tab:nemotron35-lightning-ultra-training}
\end{table*}

\begin{table*}[!tbp]
\centering
\footnotesize
\setlength{\tabcolsep}{6pt}
\begin{tabular}{lllcccc}
\toprule
Draft $\rightarrow$ target & ISL & Entropy & $T$ & 2K & 4K & 8K \\
\midrule
\multirow{2}{*}{Qwen3.5-4B $\rightarrow$ Qwen3.5-122B}
  & \multirow{2}{*}{32K} & \multirow{2}{*}{Low}
  & 0 & $+3.88\%$ & $+2.38\%$ & $+1.69\%$ \\
  & & & 1 & $+10.03\%$ & $+7.88\%$ & $+6.70\%$ \\
\bottomrule
\end{tabular}
\caption{Relative AL improvement from compressibility training on \riftada.
}
\label{tab:trainable-snapkv-adaptive-improvement}
\end{table*}

\section{Extended Related Work}
\label{app:related-work}
We review the work most directly connected to frozen prefill selection, decode-time cache management, and compressed-cache drafting.

\newparagraph{Prefill-based KV cache compression.}
A major line of work focuses on a single compression of the prompt KV cache, right before decode takes place.
SnapKV estimates prefill-token importance from attention scores produced by an observation window, often at the end of prefill~\cite{li2024snapkv}.
PyramidKV and AdaKV build SnapKV's selection paradigm while heterogeneously allocating the cache budget across layers and heads, respectively~\cite{cai2024pyramidkv,feng2025adakv}. Other static KV cache selectors change the importance signal. For example, CriticalKV scores the output perturbation caused by eviction~\cite{feng2026criticalkv}, Expected Attention models attention under a distribution of future queries~\cite{devoto2025expectedattention} and CurDKV uses query-free key and value-leverage scores~\cite{sengupta2025curdkv}.
DefensiveKV instead challenges the usual mean aggregation of importance over an observation window and introduces a worst-case-risk-aware aggregation rule. nevertheless, its eviction is still performed during prefill, leaving the retained prompt set fixed during decode~\cite{feng2025defensivekv}.
ChunkKV changes the selection unit from individual tokens to contiguous chunks~\cite{liu2025chunkkv}. Collectively, these methods primarily optimize \emph{which} entries survive compression. Our work studies a common failure mode: even a strong prefill decision can become stale as a long generated prefix changes what the model needs, and we do that through the lenses of speculative decoding.
In the context of compressibility training, KV-CAT~\cite{gelberg2026kvcat} demonstrates that training under KV sparsification can improve the post-hoc compressibility of pretrained language models, enabling additional gains through training. They consider an additional, trainable masker module, based on which sparse attention masks are built. In this work we directly consider specific KV scoring methods.

\newparagraph{Decode-time KV compression for long generation.}
Decode-time strategies continually adapt the working cache. For example, ScissorHands exploits persistence of token importance, StreamingLLM retains attention sinks and recent tokens, and H$_2$O allocates half of its cache budget to recent tokens and half to heavy-hitters identified by accumulated attention scores~\cite{liu2023scissorhands,xiao2024streamingllm,zhang2023h2o}. R-KV utilizes reasoning trace information sparsity~\cite{cai2025rkv}. QUEST proposes lightweight page-level retrieval through aggregated metrics~\cite{tang2024quest}. Its page summaries are complementary to our study of when and from which temporal region a drafter should refresh its active cache.
LongFlow derives a low-overhead importance signal from the current query for continuous eviction, whereas KARA restricts decode-time scoring to a recent window and expands selected tokens into flexible chunks~\cite{su2026longflow,han2026kara}.

\newparagraph{Compression-aware speculative decoding.}
TriForce uses a dynamically retrieved target cache as an intermediate drafter, refreshed periodically or by acceptance gating~\cite{sun2024triforce}. MagicDec studies sparse-KV self-speculation in large concurrencies, while SpecExtend and SparseSpec reuse verification attention to select KVs~\cite{sadhukhan2025magicdec,cha2025specextend,zhao2025sparsespec}. LongSpec trains a constant-context drafter and Windowed-MTP and Attention Drift study SWA-based MTP drafting and stepwise attention changes in tail-augmented drafting~\cite{yang2025longspec,valliappan2026windowedmtp,eldenk2026attentiondrift}. RIFT complements these approaches by operating directly on external or self-speculative draft caches and studying recency allocation and draft-side refresh signals under a fixed active KV cache budget.

\end{document}